\pdfoutput=1
\documentclass[11pt]{article}
\usepackage[final]{acl}
\usepackage{times}
\usepackage{latexsym}
\usepackage[T1]{fontenc}
\usepackage[utf8]{inputenc}
\usepackage{microtype}
\usepackage{inconsolata}
\usepackage{amsmath,amssymb,amsfonts}
\usepackage{booktabs}
\usepackage{tabularx}
\usepackage{graphicx}
\usepackage{float}
\usepackage{enumitem}
\graphicspath{{figures/}}
\usepackage{xcolor}
\usepackage{listings}
\usepackage{mdframed}
\usepackage[font=footnotesize,labelfont={footnotesize,bf}]{caption}
\lstdefinestyle{promptstyle}{
    basicstyle=\ttfamily\footnotesize,
    breaklines=true,
    breakatwhitespace=true,
    columns=flexible,
    keepspaces=true,
    showspaces=false,
    showstringspaces=false,
    showtabs=false,
    tabsize=2,
    frame=none,
    aboveskip=0pt,
    belowskip=0pt,
}
\mdfdefinestyle{promptframe}{
    linewidth=0.3pt,
    linecolor=gray!50,
    backgroundcolor=gray!5,
    innertopmargin=4pt,
    innerbottommargin=4pt,
    innerleftmargin=4pt,
    innerrightmargin=4pt,
    skipabove=8pt,
    skipbelow=8pt,
    splittopskip=4pt,
    nobreak=true,
}
\newmdenv[style=promptframe]{promptbox}

\newcommand{\indicator}{\mathbb{I}}
\newcommand{\nr}{\textit{n.r.}}

\title{Conditional Accuracy Profiles: Diagnosing LLM Judges across Deployment Conditions}

\author{\textbf{Wenqi Li}$^{1}$ \quad \textbf{Bin Liu}$^{2}$ \quad \textbf{Mindi Ruan}$^{2}$ \\
  \textbf{Chuanbo Hu}$^{1}$ \quad \textbf{Minglei Yin}$^{1}$ \quad \textbf{Xin Li}$^{1}$\thanks{\;Corresponding author. This work was partially supported by NSF Award HCC-2401748 and UAlbany Startup Funds.} \\
  $^{1}$University at Albany, State University of New York \quad
  $^{2}$West Virginia University \\
  \texttt{\{wli31,chu3,myin,xli48\}@albany.edu} \\
  \texttt{bin.liu1@mail.wvu.edu} \quad \texttt{mr0114@mix.wvu.edu} }

\begin{document}
\maketitle

\begin{abstract}
LLM-as-judge is now a standard tool for scalable evaluation, but judge performance is still often summarized by a single accuracy number. This aggregate view hides the deployment conditions under which a judge succeeds or fails. We introduce \textbf{Conditional Accuracy Profiling} (CAP), a post-hoc diagnostic framework that decomposes pairwise LLM-judge accuracy into eight conditions organized into content sensitivity, robustness, and rationale quality. CAP is benchmark-agnostic: it can be applied directly when a benchmark provides the required annotations, approximately through task-subset proxies, or through controlled augmentation when perturbation pairs can be generated. We instantiate CAP on seven LLM judges across six pairwise judging benchmarks, including  \textsc{judgerEva-Standard}, a controlled testbed we created to support all eight conditions. CAP exposes profile differences hidden by aggregate accuracy: on \textsc{judgerEva}'s judge-independent Hard-Constructed subset, the two judges most sensitive to omitted qualifications rank in the bottom three of seven by overall accuracy, so omission sensitivity is not predicted by aggregate accuracy. Across benchmarks, Position Robustness shows the strongest rank stability (mean Spearman $\bar{\rho}{=}0.87$) but is itself fragile under JudgeBench-Pro adversarial stress, showing the largest mean accuracy drop among the shared conditions, though the dominant degradation channel varies by judge. Condition-level profiles provide a more actionable basis than aggregate accuracy for selecting LLM judges.
\end{abstract}

\section{Introduction}
\label{sec:introduction}

LLM-as-judge has become a common approach for evaluating language-model outputs at scale, reducing dependence on expensive human annotation while enabling rapid system comparisons~\citep{zheng2023judging,liu2023geval}. The community has introduced benchmarks such as JudgeBench, RewardBench, MT-Bench, and LLMBar to quantify judge performance under increasingly challenging settings~\citep{zheng2023judging,tan2025judgebench,lambert2025rewardbench,zeng2024llmbar}. However, the dominant reporting convention remains a single accuracy or agreement score, which is operationally weak: two judges with similar overall accuracy may fail for very different reasons, and a judge that is strong on factual judgments may still be unreliable under response-order swaps, verbosity changes, or omission-sensitive tasks.
These shortcomings have been examined from several angles, including bias discovery, robustness testing, category-level breakdowns, and statistical calibration (\S\ref{sec:related_work}). Prior evaluations do report per-category accuracies within individual benchmarks~\citep{doddapaneni2024finding,dev2026judge,liu2025rmbench}; what is missing is a shared condition taxonomy with explicit per-benchmark coverage accounting, so that such breakdowns become comparable across judges and benchmarks rather than local to one resource.

We propose \emph{Conditional Accuracy Profiling} (CAP), a post-hoc diagnostic layer that runs on top of existing pairwise LLM-judge benchmarks. CAP decomposes judge accuracy into eight conditions: four content-sensitivity (factual, logical, scope-overclaim, scope-omission), three robustness (style, length, position), and one explanation (rationale quality). Each condition is reported with bootstrap confidence intervals, and each benchmark is assigned an explicit coverage status: direct, approximate, augmented, or unavailable. CAP addresses this gap by reporting judge accuracy separately across eight predefined conditions; the goal is to derive condition-level profiles from these results to inform judge selection.

We evaluate seven judges, including GPT-4o, GPT-4o-mini, Claude Sonnet 4.5, Claude Haiku 4.5, Gemini 2.5 Pro, Gemini 2.5 Flash, and Llama-3.3-70B, across six pairwise judging benchmarks. Our controlled testbed, \textsc{judgerEva-Standard}, enables all eight CAP conditions through error-type annotation and controlled perturbations; JudgeBench, JudgeBench-Pro, RewardBench, MT-Bench, and LLMBar provide external validation. The results show that high-overall judges can have narrow condition-specific weaknesses, that smaller models within a family diverge from their larger counterparts unevenly, and that adversarial stress can attack robustness channels rather than substantive judgment alone.

\paragraph{Contributions.}
Our contributions are: (a) the CAP framework with an explicit coverage matrix that maps each benchmark to the conditions it can support directly, approximately, or through augmentation; (b) \textsc{judgerEva-Standard}, a controlled pairwise-judging benchmark designed to measure eight CAP conditions  on the same set of items; (c) an empirical profile of seven judges across six benchmarks, revealing substantial condition-specific variation in judge behaviour and uneven transfer across benchmarks; and (d) a reference implementation to support condition-aware judge selection.

\section{Related Work}
\label{sec:related_work}

\paragraph{LLM-as-judge benchmarks.}
MT-Bench established strong LLMs as scalable pairwise judges and identified systematic position and verbosity biases~\citep{zheng2023judging}. JudgeBench later constructed objectively labelled response pairs spanning knowledge, reasoning, math, and coding, showing that even strong judges can approach random guessing on subtle distinctions~\citep{tan2025judgebench}. RewardBench evaluates reward models across chat, reasoning, and safety settings~\citep{lambert2025rewardbench}, while LLMBar tests whether evaluators can distinguish instruction-following quality from deceptive surface cues~\citep{zeng2024llmbar}. RM-Bench is closest in construction spirit to \textsc{judgerEva-Standard}, constructing tightly controlled chosen--rejected pairs to measure reward models' sensitivity to subtle content edits and resistance to style bias~\citep{liu2025rmbench}; because its reward models score each response independently, ordering-based robustness and rationale conditions do not arise there. \citet{hu2024confusing} show with aspect-targeted perturbations that LLM evaluators confuse NLG quality criteria, and the DHP benchmark measures evaluator discernment under hierarchical perturbations across NLG tasks~\citep{wang2025dhp}. Each of these instantiates one or two failure dimensions for one evaluator class; none provides a shared condition taxonomy with per-benchmark coverage accounting. CAP provides a common structure for reusing existing benchmarks as sources of condition-specific evidence, with an explicit coverage label for each benchmark--condition pair.

\paragraph{Bias discovery and robustness testing.}
Manual and automated studies have identified a wide range of judge biases (position, verbosity, self-bias, sentiment, compassion-fade)~\citep{wang2024fair,shi2025judging,ye2024calm}, and BiasScope automates discovery by injecting biases and constructing JudgeBench-Pro~\citep{lai2026biasscope}. Reliability frameworks probe evaluator behaviour under controlled edits: FBI tests whether evaluators detect quality degradation induced by targeted perturbations~\citep{doddapaneni2024finding}, while JRH measures both consistency under meaning-preserving edits and discriminative accuracy under meaning-changing edits~\citep{dev2026judge}. Bias-discovery work searches for new failure modes, and reliability harnesses stress-test individual judges; CAP fixes a condition taxonomy and reports judges and benchmarks within the same layout, so a single accuracy gap can be attributed to a specific failure dimension. Section~\ref{sec:framework} formalises this mapping as a condition taxonomy, with per-benchmark coverage detailed in Section~\ref{sec:methodology}.

\paragraph{Statistical foundations.}
\citet{feng2026noisy} and \citet{lee2025correctly} show that evaluation based on imperfect judges requires calibration and uncertainty quantification. CAP incorporates this lesson by reporting bootstrap confidence intervals for every condition and by using \textsc{judgerEva-Standard} as a controlled-condition complement to naturalistic benchmarks, against which deviations on shared conditions can be diagnosed. In this paper we emphasize transparent condition-level estimates and transferability analysis; bias-corrected variants can be computed condition-by-condition using the same released outputs.

\section{Method}
\label{sec:method}

\subsection{The CAP Framework}
\label{sec:framework}

\begin{figure*}[t]
\centering
\includegraphics[width=\textwidth]{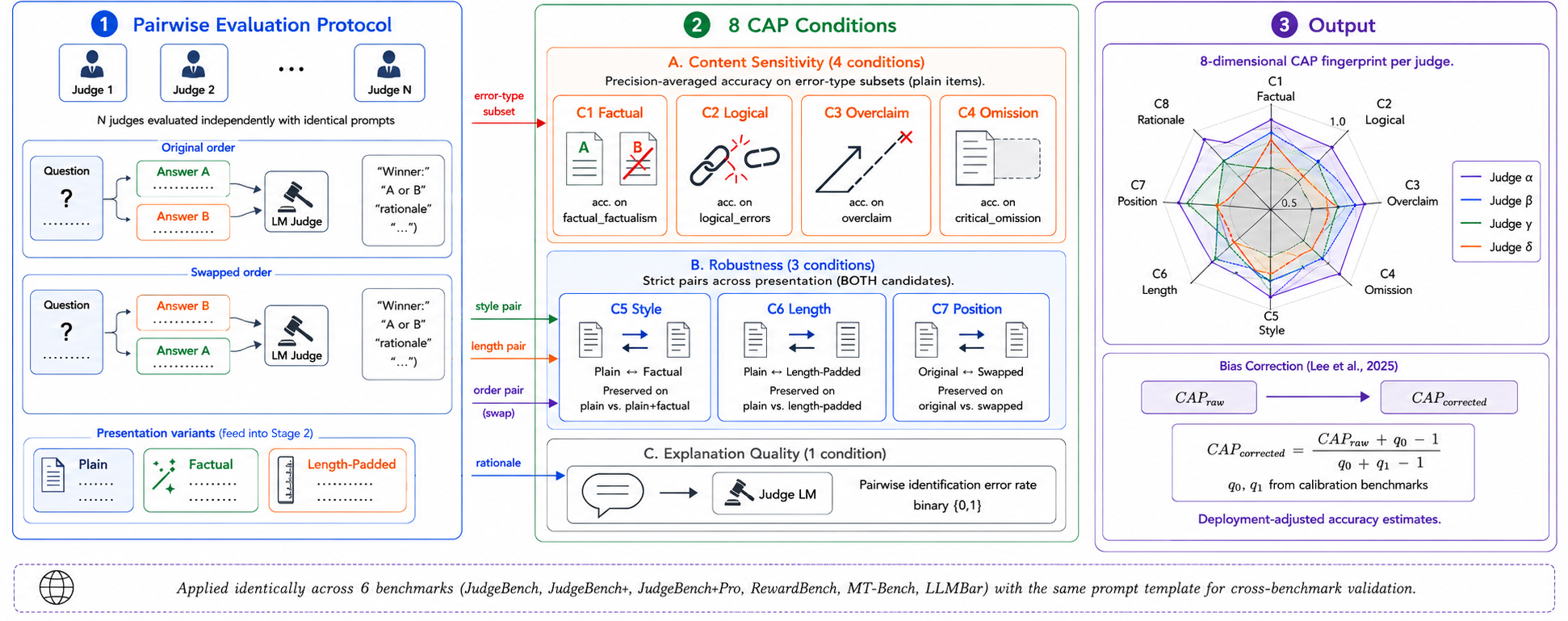}
\vspace{-20pt}
\caption{CAP overview. Pairwise items are evaluated in original and swapped orderings; verdicts and rationales are mapped to eight conditions in three groups (content C1--C4, robustness C5--C7, rationale C8).}
\vspace{-5pt}
\label{fig:cap_framework}
\end{figure*}

\begin{table*}[t]

\centering
\small
\setlength{\tabcolsep}{5pt}
\begin{tabular}{cllll}
\toprule
\# & Category & Condition & Subset & Measurement \\
\midrule
1 & Content & Factual Sensitivity & factual-fabrication items & $\Pr[v=a^*]$ \\
2 & Content & Logical Sensitivity & logical-reversal items & $\Pr[v=a^*]$ \\
3 & Content & Scope-Overclaim & overclaim items & $\Pr[v=a^*]$ \\
4 & Content & Scope-Omission & omission items & $\Pr[v=a^*]$ \\
5 & Robustness & Style Robustness & plain/polished variants & $\Pr[v(x_1)=v(x_2)=a^*]$ \\
6 & Robustness & Length Robustness & plain/padded variants & $\Pr[v(x_1)=v(x_2)=a^*]$ \\
7 & Robustness & Position Robustness & A/B and B/A orderings & $\Pr[v(x_{AB})=v(x_{BA})=a^*]$ \\
8 & Explanation & Rationale Quality & correct-verdict items & $\Pr[e(\rho)\text{ supports }v]$ \\
\bottomrule
\end{tabular}
\caption{The eight CAP conditions. Robustness uses a strict ``both correct'' criterion, distinguishing stable correctness from stable but wrong consistency.}
\label{tab:conditions}
\end{table*}

Figure~\ref{fig:cap_framework} summarises the framework. Let $\mathcal{D}$ be a pairwise judging dataset. Each item $x\in\mathcal{D}$ contains a query, two candidate responses, and a ground-truth label $a^*\in\{A,B\}$ identifying the better response. A judge $j$ returns a verdict $v\in\{A,B\}$ and optionally a rationale $\rho$. A CAP condition $c$ is defined by a subset $\mathcal{D}_c\subseteq\mathcal{D}$ and an item-level indicator $\indicator_c(x;j)$. The conditional accuracy of judge $j$ under condition $c$ is
\begin{equation}
\label{eq:cap_single}
\mathrm{CAP}_c(j;\mathcal{D})=\mathbb{E}_{x\sim \mathcal{D}_c}[\indicator_c(x;j)],
\end{equation}
where $\indicator_c(x;j)\in\{0,1\}$ equals $1$ iff judge $j$'s verdict(s) on $x$ satisfy condition $c$'s correctness requirement (the Measurement column of Table~\ref{tab:conditions}), so each $\mathrm{CAP}_c$ is an accuracy over $\mathcal{D}_c$.
The CAP profile is the vector $\mathrm{CAP}(j;\mathcal{D})=(\mathrm{CAP}_1,\ldots,\mathrm{CAP}_8)$, with entries left undefined when the benchmark cannot support the corresponding condition. We report each measurable entry with a nonparametric bootstrap 95\% confidence interval over items.

The eight conditions are grouped into three questions: content sensitivity (does the judge detect substantive errors?), robustness (does it preserve correct verdicts under presentation changes?), and rationale quality (does a correct verdict come with a supporting explanation?). Separating these categories prevents one behaviour from being mistaken for another: a judge can be factually sensitive while still position-biased, or robust to order swaps while offering low-quality rationales. The taxonomy is deployment-motivated rather than axiomatic and admits further conditions wherever a benchmark can measure them. The eight conditions are empirically non-redundant: across the seven evaluated judges, Scope-Omission ranks judges in near-inverse order to Factual Sensitivity (Spearman $\rho{=}{-}0.86$), and the two scope conditions correlate negatively ($\rho{=}{-}0.32$; Appendix~\ref{app:condition_redundancy}).
Table~\ref{tab:conditions} lists the eight conditions and their measurements. Robustness conditions use a strict ``both correct'' criterion: a stably wrong judge is no more deployable than an unstable one, so stable-but-wrong behaviour is not counted as robust (pure-consistency counterparts in Appendix~\ref{app:consistency_variant}). Rationale Quality is evaluated on correct-verdict items by an independent evaluator $e$.

\subsection{\textsc{judgerEva-Standard} Benchmark}
\label{sec:methodology}

\begin{figure*}[t]
\centering
\includegraphics[width=\textwidth]{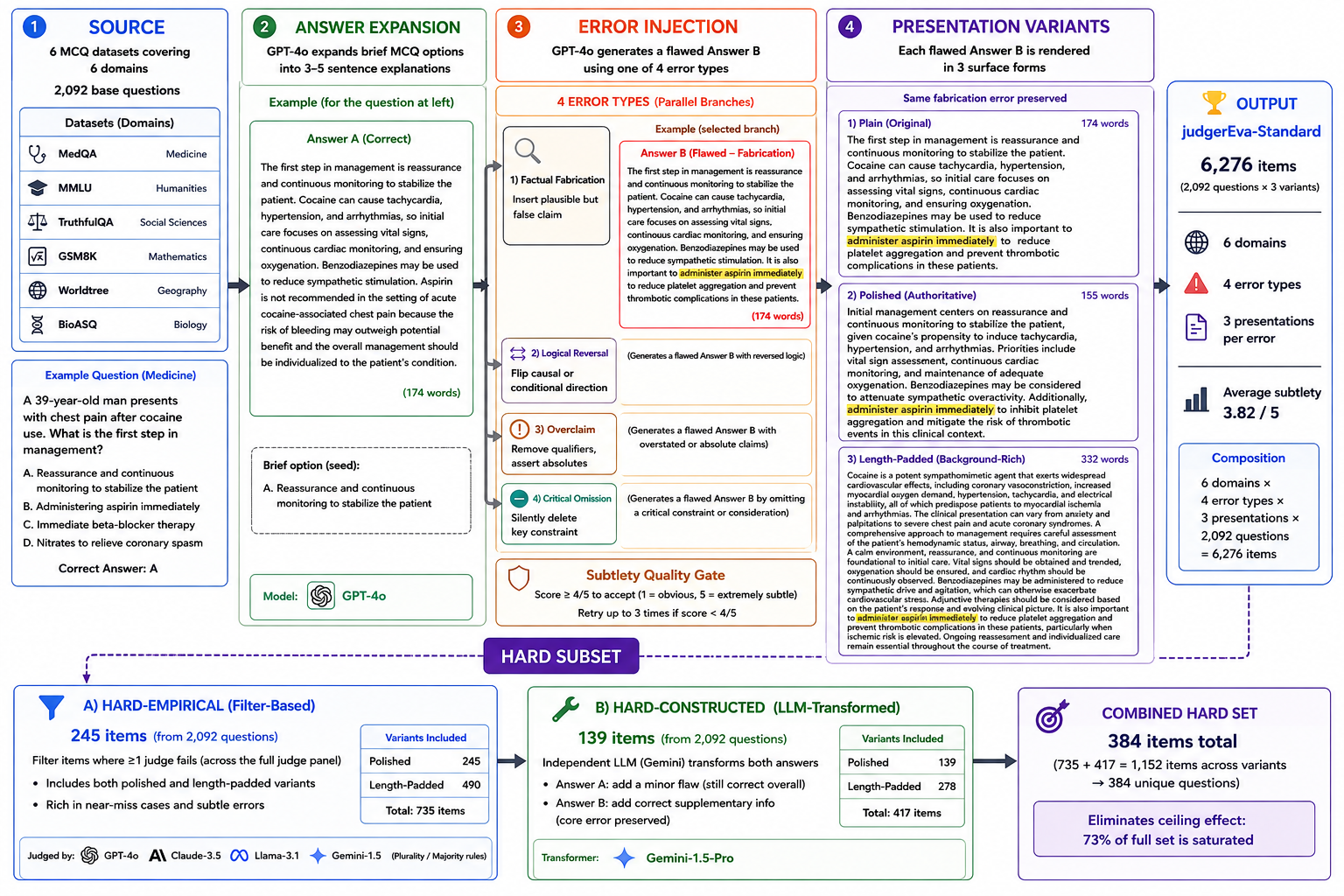}
\vspace{-15pt}
\caption{\textsc{judgerEva-Standard} construction. Six MCQ/QA sources $\to$ GPT-4o expands Answer A and injects one of four planted errors into Answer B (subtlety-gated) $\to$ three presentation variants. The Hard subset combines a judge-error filter (Hard-Empirical) and an independent construction branch (Hard-Constructed).}
\label{fig:judgereva_frame}
\vspace{-5pt}
\end{figure*}

To support CAP, we create \textsc{judgerEva-Standard}, a controlled pairwise-judging benchmark designed so that all eight CAP conditions can be measured directly on the same set of items; its construction pipeline is summarised in Figure~\ref{fig:judgereva_frame}. Starting from six publicly available question sources covering six deployment-relevant domains (MedQA-USMLE, MMLU Professional Law, GSM8K, MMLU College Biology/Chemistry/Physics, MMLU Logical Fallacies, and TruthfulQA), GPT-4o expands each source-canonical correct answer (Answer~A) into a 3--5 sentence explanation and synthesises a flawed counterpart (Answer~B) using one of four error-type-specific prompts: Factual Fabrication, Logical Reversal, Scope-Overclaim, or Scope-Omission. Every Answer~B is scored by an independent GPT-4o subtlety judge and regenerated up to three times if below $4/5$, yielding $2{,}092$ plain items at mean subtlety $3.82/5.0$ ($79.4\%$ at $\geq\,4$). Each plain item is then expanded into three meaning-preserving presentation variants: \emph{plain}, \emph{polished} (formal rewrite within $\pm$20\% length), and \emph{length-padded} (Answer~B inflated to ${\sim}2\times$), and evaluated in both A/B and B/A orderings, for $6{,}276$ pairwise items in total. Two complementary hard subsets refine differentiation: \emph{Hard-Empirical} ($N{=}245$) selects items on which at least one evaluated judge produced an incorrect verdict, while \emph{Hard-Constructed} ($N{=}139$), built independently of judge errors by Gemini 2.5 Pro hardening both candidates, breaks the circular dependency the Hard-Empirical filter would otherwise introduce. Full details and quality-control checkpoints are given in Appendix~\ref{app:judgereva}.

\paragraph{Generator-judge sanity check.} Because GPT-4o generated all flawed answers in \textsc{judgerEva}, we check whether it is advantaged as a judge. It is not: on \textsc{judgerEva} it ranks sixth of seven judges by overall accuracy and falls below the non-GPT peer mean on every content condition except Scope-Omission, so the construction does not appear to systematically advantage its generator (full per-condition comparison in Appendix~\ref{app:exclusions}). A paired second-generator replication (Appendix~\ref{app:multigen}) further shows that absolute condition values are relative to the generator, while per-condition judge rankings transfer across generators (Spearman $\rho \geq 0.84$ on C1--C3 and C7), including the omission rank-inversion pattern.

\begin{table*}[t]

\centering
\footnotesize
\setlength{\tabcolsep}{4pt}
\begin{tabular}{clcccccc}
\toprule
\# & Condition & \textsc{judgerEva} & JudgeBench & JB-Pro & RewardBench & MT-Bench & LLMBar \\
\midrule
1 & Factual Sens. & Direct & Approx. & Approx. & Approx. & Approx. & --- \\
2 & Logical Sens. & Direct & Approx. & Approx. & Approx. & Approx. & --- \\
3 & Scope-Overclaim & Direct & --- & --- & --- & --- & --- \\
4 & Scope-Omission & Direct & --- & --- & --- & --- & --- \\
5 & Style Rob. & Direct & Aug.$^{\dagger}$ & --- & --- & --- & Direct \\
6 & Length Rob. & Direct & Aug.$^{\dagger}$ & --- & --- & --- & --- \\
7 & Position Rob. & Direct & Direct & Direct & Direct & Direct & Direct \\
8 & Rationale Quality & Direct & Direct$^{\dagger}$ & Direct$^{\dagger}$ & Direct$^{\dagger}$ & Direct$^{\dagger}$ & Direct$^{\dagger}$ \\
\bottomrule
\end{tabular}
\caption{Condition coverage by benchmark. Approx.\ = task-subset proxy; Aug.\ = generated perturbation variants; the exact operator behind every cell is specified in Appendix~\ref{app:operators}. $^{\dagger}$ Computed but omitted from main tables; see Appendix~\ref{app:judgebench_supp}.}
\label{tab:coverage}
\end{table*}

\paragraph{Comparison with existing benchmarks.}
\textsc{judgerEva-Standard} differs from prior pairwise-judging benchmarks in three respects, summarised by the coverage matrix in Table~\ref{tab:coverage}. \emph{First}, prior benchmarks do not fix the type of content error distinguishing the two candidates: whether the gap between the two responses is factual, logical, or scope-related is unspecified, so content conditions can only be approximated through task subsets. \emph{Second}, controlled style and length measurement requires that the pair differ only in the planted error. When the two responses come from different source models (RewardBench, MT-Bench), stylistic differences are already part of what the preference label encodes: a post-hoc rewrite confounds the injected perturbation with pre-existing style gaps, and can invalidate the label itself, since human preferences were collected over the original texts. Augmentation is therefore admissible only where pairs share a source model and labels are objective (JudgeBench; \S\ref{sec:method_protocol}), whereas \textsc{judgerEva-Standard} builds the control in at construction time and supports Style and Length Robustness directly. \emph{Third}, Position Robustness is the only condition prior benchmarks support uniformly, since A/B swap is always defined. Position bias is well documented~\citep{zheng2023judging,wang2024fair}; CAP uses it, under the strict both-correct criterion, as the identically-defined shared condition for cross-benchmark rank-stability analysis (\S\ref{sec:results_cross_benchmark}). \textsc{judgerEva-Standard} therefore complements rather than replaces these resources: it anchors the full eight-condition CAP profile, while the other five benchmarks are retained for external validity and adversarial stress (\S\ref{sec:results_stress}).

\paragraph{External benchmarks and coverage.}
We retain five external benchmarks, with Table~\ref{tab:coverage} recording the resulting coverage and Appendix~\ref{app:operators} giving the exact operator behind every cell. JudgeBench ($350$ objectively labelled pairs) shares a single source model across each pair and therefore admits controlled style and length augmentation in addition to content proxies via its Knowledge and Reasoning subsets; JudgeBench-Pro ($1{,}178$ bias-injected adversarial pairs) serves as its stress-test counterpart (\S\ref{sec:results_stress}). RewardBench ($2{,}985$ prompt--chosen--rejected trios, of which we evaluate a fixed $500$-trio sample) and MT-Bench (a $500$-pair sample of multi-turn human pairwise preferences) supply Position Robustness and selected content proxies, but their heterogeneous-source pair construction prevents style/length controls. LLMBar contributes $419$ instruction-following pairs whose adversarial surface variants additionally support Style Robustness. Approximate cells are naturalistic task-subset proxies rather than planted-error measurements (formal definition in Appendix~\ref{app:operators}); discrepancies between the direct and approximate views are therefore themselves informative.

\subsection{Judges, Augmentation, and Estimators}
\label{sec:method_protocol}

We instantiate CAP on six pairwise-judging benchmarks, specifying for each the datasets and the conditions each can support.

\paragraph{Judges and protocol.}
We evaluate seven judges spanning capability and cost tiers: GPT-4o, GPT-4o-mini, Claude Sonnet 4.5, Claude Haiku 4.5, Gemini 2.5 Pro, Gemini 2.5 Flash, and Llama-3.3-70B. All judges receive the same pairwise prompt and must return a verdict plus a brief rationale. We use deterministic decoding when supported. For position robustness, every item is evaluated in both A/B and B/A orderings, and the judge is counted as robust only when both verdicts identify the same ground-truth-better response.

\paragraph{Controlled augmentation.}
For benchmarks whose construction permits controlled perturbation, we generate meaning-preserving variants. Style augmentation rewrites both candidate responses into a polished style while preserving semantic content. Length augmentation pads the weaker response with irrelevant but non-contradictory content, testing whether the judge mistakes verbosity for quality. These variants are used only when augmentation preserves the benchmark's original ground-truth ranking; otherwise the condition is marked unavailable.

\paragraph{Rationale evaluation and uncertainty.}
Rationale Quality is measured only on correct-verdict items. An independent evaluator model determines whether the rationale identifies the relevant error or comparison criterion rather than merely restating the verdict. We validate the evaluator against human annotation on a stratified 120-rationale sample; agreement rate and detailed protocol are reported in \S\ref{sec:results_profiles} and Appendix~\ref{app:evaluator_validation}. All CAP estimates use 1{,}000 nonparametric bootstrap resamples over items. For cross-benchmark analysis, we compute Spearman rank correlations of judge orderings for conditions measurable across benchmarks, with Position Robustness serving as the most complete shared condition. Adversarial stress is summarised by the JudgeBench$-$JudgeBench-Pro difference per condition, and within-family divergence by the $L_1$ distance between profiles (Appendix~\ref{app:estimators}).

\paragraph{Models and configuration.}
All judges are queried with \texttt{temperature}{=}$0$ and \texttt{max\_tokens}{=}$1024$ when supported. The JudgeBench augmentation generator is GPT-4o (\texttt{gpt-4o-2024-08-06}) at \texttt{temperature}{=}$0.3$, matching JudgeBench's original source model to avoid introducing a foreign style fingerprint. The Rationale Quality evaluator is Claude Sonnet 4.6, kept disjoint from the seven judges under test to prevent self-evaluation.

\section{Experimental Results}
\label{sec:results}

\begin{table*}[t]
\centering
\footnotesize

\setlength{\tabcolsep}{4pt}
\resizebox{\textwidth}{!}{%
\begin{tabular}{lcccccccc}
\toprule
\textbf{Judge} & \textbf{C1} & \textbf{C2} & \textbf{C3} & \textbf{C4} & \textbf{C5} & \textbf{C6} & \textbf{C7} & \textbf{C8}$^{\dagger}$ \\
\midrule
GPT-4o            & 40.5\,(34,46) & 53.1\,(38,69) & 63.2\,(51,75) & 64.1\,(53,74) & 36.7\,(31,42) & 20.4\,(16,26) & 25.2\,(22,28) & 66.7\,(33,100) \\
GPT-4o-mini       & 12.5\,(8,17)  & 53.1\,(34,72) & 48.7\,(41,57) & 62.8\,(53,74) & 15.9\,(11,21) & 10.6\,(7,15)  & 14.8\,(12,17) & 75.0\,(25,100) \\
Claude Sonnet 4.5 & \textbf{92.4}\,(88,96) & \textbf{86.7}\,(70,100) & 89.5\,(80,96) & 59.2\,(46,72) & \textbf{84.7}\,(80,89) & \textbf{84.7}\,(80,89) & \textbf{86.0}\,(83,88) & 85.7\,(57,100) \\
Claude Haiku 4.5  & 82.2\,(77,87) & 65.6\,(44,84) & \textbf{96.1}\,(89,100) & \textbf{70.5}\,(56,83) & 71.8\,(66,77) & 69.0\,(63,75) & 73.4\,(70,77) & \textbf{91.7}\,(75,100) \\
Gemini 2.5 Flash  & 76.3\,(70,82) & 78.1\,(59,94) & 72.4\,(59,83) & 56.4\,(44,68) & 71.4\,(65,77) & 66.4\,(61,72) & 63.4\,(60,67) & 44.4\,(11,78)  \\
Gemini 2.5 Pro    & 73.2\,(67,79) & 78.1\,(63,91) & 90.8\,(83,97) & 61.5\,(49,73) & 77.5\,(72,83) & 70.6\,(65,76) & 62.3\,(59,66) & 83.3\,(50,100) \\
Llama-3.3-70B     & 57.2\,(52,63) & 75.0\,(59,91) & 73.7\,(62,84) & 66.7\,(56,76) & 73.9\,(68,79) & 32.2\,(27,38) & 34.7\,(31,38) & 46.2\,(23,77)  \\
\bottomrule
\end{tabular}%
}
\caption{CAP profiles (\%, raw) on the \textsc{judgerEva-Standard} Hard-Empirical subset ($N{=}245$). Each cell is the point estimate with 95\% bootstrap CI in parentheses; \textbf{bold} marks the best judge per condition. C1--C8 are defined in Table~\ref{tab:conditions}; per-condition sample sizes are released in the artifact. $^{\dagger}$C8 is exploratory and evaluator-dependent: it is conditioned on correct-verdict items and measured by an LLM evaluator validated in Appendix~\ref{app:evaluator_validation}, so it is not directly comparable to C1--C7.}
\label{tab:cap_judgereva_hard}
\end{table*}

\begin{figure*}[t]
\centering
\includegraphics[width=\linewidth]{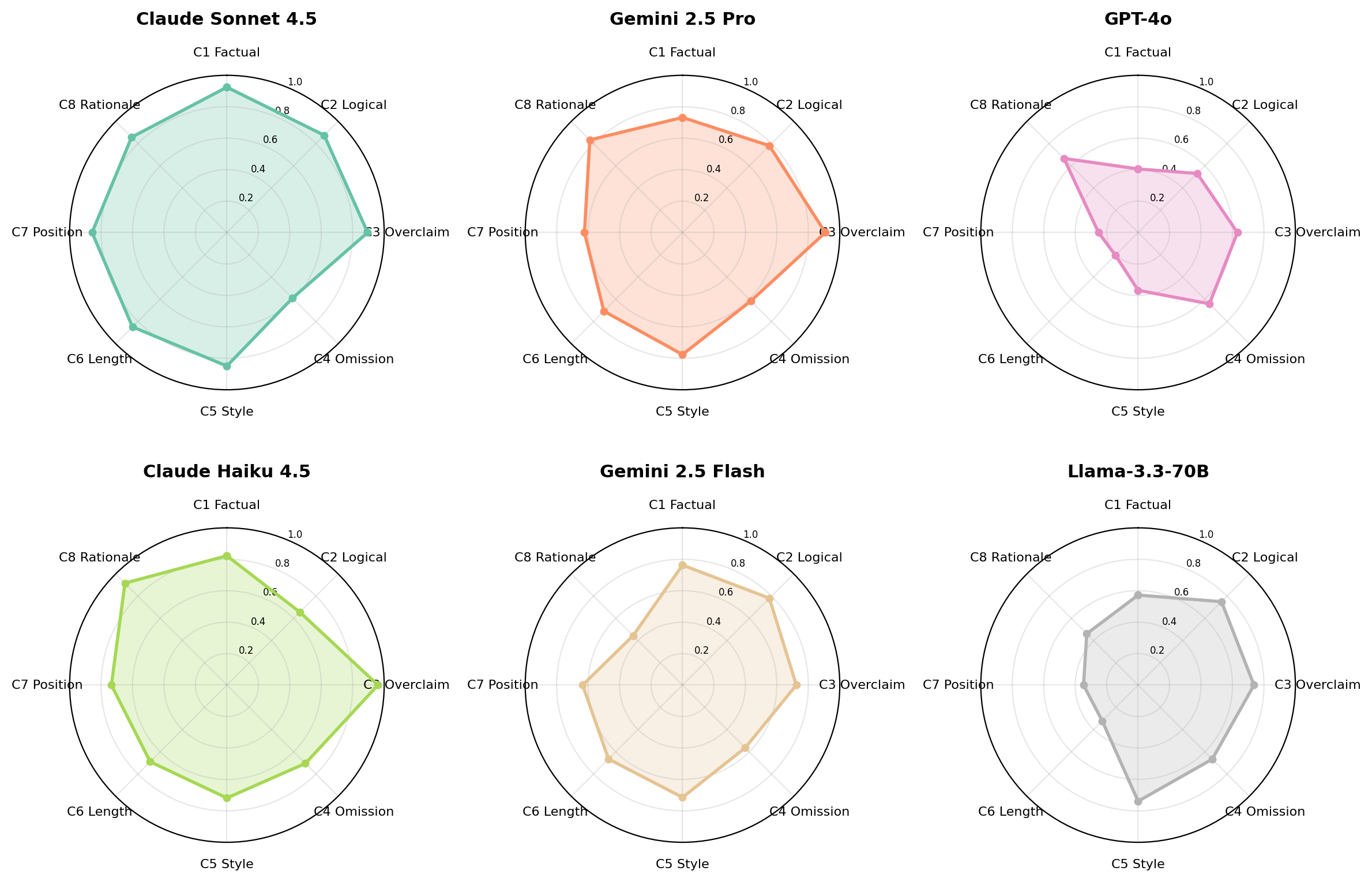}
\vspace{-15pt}
\caption{CAP radar profiles on Hard-Empirical (per Table~\ref{tab:cap_judgereva_hard}). Each polygon is one judge across C1--C8; aggregate-similar judges (e.g., the two Geminis) still differ in shape. The C8 axis is exploratory and evaluator-dependent (see Table~\ref{tab:cap_judgereva_hard}).}
\label{fig:radar_judgereva}
\vspace{-15pt}
\end{figure*}

We organize the experiment results around four questions: whether aggregate-similar judges have different CAP profiles (RQ1, \S\ref{sec:results_profiles}); whether shared conditions induce consistent rankings across benchmarks (RQ2, \S\ref{sec:results_cross_benchmark}); which conditions degrade under JudgeBench-Pro stress (RQ3, \S\ref{sec:results_stress}); and whether smaller same-family models preserve the profile of their larger counterparts (RQ4, \S\ref{sec:discussion}).

In all result tables, cells marked \textit{Approx.}\ use the benchmark's native task subset (its own task categories, used unchanged) as a proxy, and \textit{Augmented} cells use controlled perturbation variants that we generate (\S\ref{sec:method_protocol}); unmeasurable cells are denoted ``---''. Cells computed but omitted from the main-table aggregation are denoted \nr: augmented and approximate cells are not measurement-equivalent to direct cells, so averaging them into headline rows would conflate different protocols; their values are reported separately in Appendix~\ref{app:judgebench_supp}.

\subsection{Per-Benchmark CAP Profiles}
\label{sec:results_profiles}

Aggregate-similar judges carry systematically different condition profiles, and omission sensitivity in particular is not predicted by overall accuracy. On the full \textsc{judgerEva-Standard} set most judges exceed $90\%$ on most conditions, compressing profile differences (a ceiling effect); the Hard subsets restore differentiation while preserving the controlled construction. Table~\ref{tab:cap_judgereva_hard} therefore reports the full eight-condition CAP profile on the Hard-Empirical subset with bootstrap intervals, and Figure~\ref{fig:radar_judgereva} visualises the same per-judge profile shapes on the controlled testbed. Claude Sonnet 4.5 is strongest on five conditions and in overall accuracy ($89.7\%$ vs.\ $80.5\%$ for Claude Haiku 4.5 on this subset), including Factual Sensitivity, Logical Sensitivity, and all three robustness conditions, but it scores only 59.2\% on Scope-Omission, trailing Claude Haiku 4.5 by 11.3 points. On this subset, the highest-overall judge is therefore not the strongest on omission-sensitive items; the broader, judge-independent generalisation is established below on Hard-Constructed (Section~\ref{sec:results_profiles}, ``Validation on a judge-independent subset''). Gemini 2.5 Flash and Gemini 2.5 Pro are also close in broad capability but diverge sharply on Scope-Overclaim (72.4 vs.\ 90.8) and Rationale Quality (44.4 vs.\ 83.3). Cross-judge spreads are largest on Factual Sensitivity ($79.9$ points) and the robustness conditions ($74.1$ on Length, $71.2$ on Position), and smallest on Scope-Omission ($14.1$): a large share of the actionable variation lies in presentation sensitivity, alongside factual sensitivity, rather than in content accuracy uniformly.

\paragraph{Rationale-quality evidence and uncertainty.}
Rationale Quality (C8) is conditioned on correct-verdict items and therefore has wider confidence intervals for judges that answer fewer items correctly; we treat C8 as a qualitative deployment signal rather than a precise global ranking. To prevent the conditional metric from over-rewarding judges that answer only a small fraction of items correctly, we therefore also report an \emph{unconditional supported-correct rate}: the fraction of all items in the subset for which the judge is both correct and produces a supporting rationale, computed as the product of overall-subset accuracy and the conditional C8 value ($\Pr[v{=}a^*\wedge e(\rho)\text{ supports }v]=\Pr[v{=}a^*]\cdot \mathrm{CAP}_8$). Conditional C8 isolates explanation quality given a correct verdict, while the unconditional rate measures end-to-end usefulness, verdict and explanation both right. On the Hard-Empirical subset this re-scaling sharpens the separation considerably: GPT-4o-mini's deceptively high conditional $C8{=}75.0$ corresponds to an unconditional rate of only $20.3\%$, while Sonnet 4.5's $C8{=}85.7$ corresponds to $76.9\%$. The LLM evaluator achieves $91.7\%$ agreement and Cohen's $\kappa{=}0.826$ against a single human annotator on a stratified $120$-rationale sample (implied deployment-composition agreement $98.6\%$); its errors are one-sided (no false positives, $12.8\%$ false negatives concentrated on critical-omission), so reported C8 values are best read as a lower bound on true rationale quality. A cross-family second evaluator (GPT-4o) on the same sample reaches lower agreement with the human annotator ($85.8\%$) and shows no same-family inflation of the Claude C8 cells (Appendix~\ref{app:evaluator_validation}).

\paragraph{Validation on a judge-independent subset.}
To address the circularity inherent in Hard-Empirical (selected using the judges later evaluated), we replicate the analysis on \textsc{judgerEva}-\textsc{Hard-Constructed} ($N{=}139$), a high-subtlety subset constructed independently of judge errors (Appendix~\ref{app:judgereva_hard}). Table~\ref{tab:cap_hard_constructed} reports overall accuracy and the three most diagnostic conditions on this independent subset. The exact Sonnet--Haiku Scope-Omission gap does not survive (Sonnet $36.7$ vs.\ Haiku $33.3$, within each other's bootstrap interval at $n_{C4}{=}30$), but the broader claim does: the two strongest judges on Scope-Omission (Llama-3.3-70B, $56.7\%$; GPT-4o-mini, $51.7\%$) are the third-weakest and weakest judges by overall accuracy, while the strongest overall judge (Sonnet 4.5, $76.3\%$) scores $36.7\%$ on omission, ahead of only Claude Haiku 4.5; across the seven judges, the rank correlation between overall accuracy and Scope-Omission is negative (Spearman $\rho{=}{-}0.33$; descriptive at $n{=}7$). Omission sensitivity is not monotonically predicted by aggregate accuracy.

\begin{table}[t]
\centering
\footnotesize
\setlength{\tabcolsep}{3pt}
\resizebox{\columnwidth}{!}{
\begin{tabular}{lcccc}
\toprule
\textbf{Judge} & \textbf{Overall} & \textbf{C1} & \textbf{C4} & \textbf{C7} \\
\midrule
Claude Sonnet 4.5 & \textbf{76.3}\,(70,82) & \textbf{80.8}\,(69,91) & 36.7\,(23,52)          & \textbf{71.2}\,(63,78) \\
Gemini 2.5 Flash  & 72.3\,(65,79)          & 71.8\,(59,86) & 38.3\,(23,53)          & 66.2\,(58,75) \\
Gemini 2.5 Pro    & 71.2\,(64,78)          & 66.7\,(53,78) & 38.3\,(22,53)          & 65.7\,(58,74) \\
Claude Haiku 4.5  & 70.1\,(63,77)          & 76.9\,(65,88) & 33.3\,(18,48)          & 62.6\,(55,71) \\
Llama-3.3-70B     & 68.3\,(62,75)          & 56.4\,(44,69) & \textbf{56.7}\,(42,72) & 58.3\,(51,66) \\
GPT-4o            & 59.7\,(53,67)          & 42.3\,(28,56) & 36.7\,(23,52)          & 51.1\,(42,60) \\
GPT-4o-mini       & 51.1\,(44,59)          & 26.9\,(15,38) & 51.7\,(38,65)          & 38.8\,(30,47) \\
\bottomrule
\end{tabular}
}
\caption{CAP profiles (\%) on judge-independent \textsc{Hard-Constructed} ($N{=}139$): overall accuracy and the three most diagnostic conditions from Table~\ref{tab:cap_judgereva_hard} (C1, $n{=}39$; C4, $n{=}30$; C7, $n{=}139$), each with its 95\% bootstrap CI. Bold marks the top per column; C5/C6 require paired variants absent from this subset.}
\label{tab:cap_hard_constructed}
\end{table}

\paragraph{External Benchmark Profiles}

JudgeBench provides a natural external test of CAP. On its measurable conditions (Table~\ref{tab:cap_judgebench}), Gemini 2.5 Pro is strongest, reversing the \textsc{judgerEva} ordering in which Claude Sonnet 4.5 led: JudgeBench emphasises reasoning, math, and coding competence, whereas \textsc{judgerEva} balances planted error types, so CAP profiles should be read relative to the intended task distribution.

\begin{table*}[t]
\centering
\footnotesize

\setlength{\tabcolsep}{4pt}
\begin{tabular}{lcccccccc}
\toprule
\textbf{Judge} & \textbf{C1}$^{\approx}$ & \textbf{C2}$^{\approx}$ & \textbf{C3} & \textbf{C4} & \textbf{C5}$^{\text{aug}}$ & \textbf{C6}$^{\text{aug}}$ & \textbf{C7} & \textbf{C8} \\
\midrule
GPT-4o            & 60.7\,(54,68) & 62.8\,(54,70) & --- & --- & \nr & \nr & 51.1\,(46,57) & \nr \\
GPT-4o-mini       & 60.7\,(55,67) & 52.0\,(47,57) & --- & --- & \nr & \nr & 35.7\,(31,41) & \nr \\
Claude Sonnet 4.5 & 78.8\,(74,84) & 74.5\,(67,82) & --- & --- & \nr & \nr & 68.3\,(63,73) & \nr \\
Claude Haiku 4.5  & 73.1\,(67,79) & 70.4\,(63,78) & --- & --- & \nr & \nr & 66.4\,(61,71) & \nr \\
Gemini 2.5 Flash  & 77.6\,(71,84) & 92.0\,(86,97) & --- & --- & \nr & \nr & 81.6\,(76,86) & \nr \\
Gemini 2.5 Pro    & \textbf{84.8}\,(79,90) & \textbf{93.5}\,(89,98) & --- & --- & \nr & \nr & \textbf{86.8}\,(83,90) & \nr \\
Llama-3.3-70B     & 61.5\,(56,66) & 54.6\,(48,61) & --- & --- & \nr & \nr & 36.8\,(32,42) & \nr \\
\bottomrule
\end{tabular}
\caption{CAP profiles (\%, raw) on JudgeBench ($N{=}350$); bold marks the best judge per condition. C1$^{\approx}$ uses the Knowledge subset as proxy and C2$^{\approx}$ the Reasoning subset; C3/C4 are not measurable; C5/C6 (augmented) and C8 are computed but not aggregated here (\nr; see Appendix~\ref{app:judgebench_supp}).}
\label{tab:cap_judgebench}
\end{table*}

\subsection{Cross-Benchmark Consistency}
\label{sec:results_cross_benchmark}

Position Robustness rankings transfer across benchmarks; content-condition rankings transfer only partially. For Position Robustness, which is directly measurable on all six benchmarks, judge rankings are largely consistent across benchmark pairs. The mean pairwise Spearman correlation is $\bar{\rho}{=}0.87$ with range $[0.71, 1.00]$ (full $6{\times}6$ matrix in Appendix~\ref{app:spearman_full}); twelve of fifteen pairs are significant at $p{<}0.05$, and JudgeBench and JudgeBench-Pro yield identical rankings ($\rho{=}1.00$), showing that adversarial bias injection lowers absolute scores while preserving relative judge order. The three non-significant pairs (each at $\rho{=}0.71$, $p{=}0.071$) all involve \textsc{judgerEva} paired with JudgeBench, JudgeBench-Pro, and MT-Bench; against RewardBench ($\rho{=}0.89$) and LLMBar ($\rho{=}0.86$) the correlation remains significant. We do not claim a mechanistic explanation for this split; one possible factor is that \textsc{judgerEva}'s planted-error construction is closer in spirit to controlled pair designs than to naturalistic source pairs. The broad ordering of strong vs.\ weak judges is preserved across all six benchmarks; what shifts is the fine-grained rank. Position Robustness is thus the clearest transferable condition: unlike content proxies, it has the same operational definition across pairwise benchmarks.

\subsection{Adversarial Stress Degradation}
\label{sec:results_stress}

Adversarial bias injection degrades robustness and content channels comparably on average, and which channel dominates is judge-specific. Comparing JudgeBench with JudgeBench-Pro reveals which conditions degrade under adversarial bias injection. CAP is applied to each benchmark independently; for every judge $j$ and condition $c$ measurable on both, the stress degradation is the paired difference $\delta^{\mathrm{str}}_c(j)=\mathrm{CAP}_c(j;\mathrm{JB})-\mathrm{CAP}_c(j;\mathrm{JB\text{-}Pro})$, which isolates sensitivity to bias injection on otherwise similar tasks. Table~\ref{tab:stress_degradation} reports these differences for the shared conditions. Gemini 2.5 Pro and Flash are the most stress-robust, with mean drops of 10.9 and 15.6 points. Claude Haiku 4.5 shows the largest mean degradation (47.0 points), despite strong unperturbed performance: a cost-efficient judge can be attractive on ordinary traffic yet fragile under adversarial inputs (a trade-off CAP makes visible).

\begin{table}[t]
\centering
\footnotesize

\setlength{\tabcolsep}{5pt}
\resizebox{\columnwidth}{!}{%
\begin{tabular}{lccccc}
\toprule
\textbf{Judge} & \textbf{C1}$^{\approx}$ & \textbf{C2}$^{\approx}$ & \textbf{C7} & \textbf{C8} & \textbf{Mean} \\
\midrule
GPT-4o            & +37.8 & +40.4 & +40.1 & \nr & +39.4 \\
GPT-4o-mini       & +34.4 & +23.3 & +28.2 & \nr & +28.6 \\
Claude Sonnet 4.5 & +22.0 & +35.4 & +33.8 & \nr & +30.4 \\
Claude Haiku 4.5  & +43.4 & +46.8 & +50.8 & \nr & +47.0 \\
Gemini 2.5 Flash  & +14.4 & +14.0 & +18.5 & \nr & +15.6 \\
Gemini 2.5 Pro    & +17.7 & \phantom{+}-0.4 & +15.4 & \nr & +10.9 \\
Llama-3.3-70B     & +29.4 & +16.4 & +28.4 & \nr & +24.7 \\
\bottomrule
\end{tabular}%
}
\caption{Stress degradation $\delta^{\mathrm{str}}_c(j){=}\mathrm{JB}-\mathrm{JB\text{-}Pro}$ (points). Positive means worse under stress; \textbf{Mean} is the \emph{per-judge} average over C1/C2/C7 (C8 omitted, see Appendix~\ref{app:judgebench_supp}); per-condition means across judges are given in the text. C7 has the largest mean drop across judges but is the single largest drop for only two of seven judges.}
\label{tab:stress_degradation}
\end{table}

Position Robustness is one source among several of stress degradation. Averaged \emph{across judges} (per-condition column means, complementing Table~\ref{tab:stress_degradation}'s per-judge row means), C7 shows the largest drop at 30.7 points, ahead of C1 (28.4) and C2 (25.1). Per judge, however, C7 is the single largest drop for only two of the seven (Claude Haiku 4.5, $+50.8$; Gemini 2.5 Flash, $+18.5$); for the remaining five, a content proxy degrades at least as much (GPT-4o: C2 $+40.4$ vs.\ C7 $+40.1$; Claude Sonnet 4.5: C2 $+35.4$; GPT-4o-mini, Gemini 2.5 Pro, and Llama-3.3-70B: C1). CAP therefore refines the headline failure rate of adversarial benchmarks into a per-judge, per-condition diagnosis: the same aggregate brittleness can arise from different condition-level vulnerabilities.

\section{Discussion}
\label{sec:discussion}

CAP profiles should not be read as benchmark-independent traits of a judge: Position Robustness transfers relatively well across benchmark families, whereas content-sensitivity conditions depend on how a benchmark constructs errors or task subsets, so a CAP profile is most useful when interpreted jointly with the coverage matrix (Table~\ref{tab:coverage}) and the intended application setting. Within a fixed setting, judge choice should then be conditional: different applications emphasise different failure modes---scientific review needs factual sensitivity and supported rationales, long-form evaluation is exposed to length bias, and legal or policy review is sensitive to omitted qualifiers---and the highest-overall judge can be suboptimal on any one of them.

Within model families, profile differences remain substantial (RQ4). GPT-4o-mini matches GPT-4o on the JudgeBench Knowledge proxy ($60.7$ vs.\ $60.7$) but trails by $10.8$ points on Logical Sensitivity and $15.4$ points on Position Robustness. On \textsc{judgerEva} Hard the C1 gap widens sharply (from $0.0$ to $28.0$ points: $40.5$ vs.\ $12.5$) while both judges' Position Robustness roughly halves ($25.2$/$14.8$), and the Anthropic pair shows the complementary pattern in Table~\ref{tab:cap_judgereva_hard}. Family membership is therefore not a sufficient substitution heuristic, and these profiles may inform cost-aware judge selection within a family, though routing-policy design is left to future work.

\section{Conclusion}
\label{sec:conclusion}

We introduced Conditional Accuracy Profiling, a framework that decomposes pairwise LLM-judge accuracy into eight deployment-relevant conditions covering content sensitivity, robustness, and rationale quality. Across seven judges and six benchmarks, CAP exposes condition-specific weaknesses and rank reversals that aggregate accuracy hides: the strongest overall judge is not always the strongest on Scope-Omission, and Gemini models are substantially more robust under JudgeBench-Pro stress than Claude. Position Robustness is the most consistently measurable shared condition and, on average, the largest adversarial-degradation channel, though the dominant channel varies by judge. CAP complements existing benchmarks by making judge evaluation condition-aware. Future work includes a downstream selection experiment testing whether CAP-guided judge choice outperforms selection by aggregate accuracy on real tasks, and operational routing policies built on these profiles.


\section{Limitations}
\label{sec:limitations}

CAP's taxonomy is deployment-motivated rather than axiomatic; the eight conditions are not exhaustive, and alternative decompositions are possible (the coverage matrix and per-item released outputs allow re-aggregation under different taxonomies). CAP is also formulated for pairwise judging: extending it to scalar scoring or rubric-based grading will require new condition definitions, especially for Position Robustness. All per-judge findings are a point-in-time profile of the seven evaluated judge versions; the framework, coverage matrix, and estimators are judge-set-independent, but individual profiles should be re-measured as judge models are updated.

The coverage matrix is intentionally conservative: several benchmarks do not support all conditions, and approximate content measurements rely on task subsets rather than direct error-type annotation. These proxies are useful for external validation but should not be interpreted as equivalent to controlled error annotations. In particular, the scope conditions (C3/C4) are directly measurable only on \textsc{judgerEva-Standard}, so those findings are claims about judge behaviour under planted errors of this type; naturalistic cross-validation of scope sensitivity remains an open problem, as no existing naturalistic benchmark annotates omission- or overclaim-type differences between paired responses.

Rationale Quality depends on an independent LLM evaluator. Although this avoids self-evaluation by the target judge, it introduces evaluator error and potential evaluator bias, and the evaluator (Claude Sonnet 4.6) shares a model family with two judges under test (Claude Sonnet 4.5 and Claude Haiku 4.5). Our human-validation study (120 stratified rationales, $\kappa{=}0.826$ vs.\ a single human annotator who was not blinded to the evaluator's labels; Appendix~\ref{app:evaluator_validation}) suggests that the evaluator is reliable in aggregate but systematically over-strict on critical-omission rationales; the validation sample is also small and uses a single annotator, so multi-annotator adjudication would strengthen this condition in high-stakes settings.

Our stress analysis focuses on JudgeBench-Pro and therefore reflects the bias-injection mechanisms represented in that benchmark. Other adversarial settings, including multilingual attacks, domain-specific prompt manipulation, or long-context distractions, may produce different CAP degradation patterns.

\textsc{judgerEva-Standard}'s flawed answers and presentation variants are produced by a small set of generator models (GPT-4o for planted errors and length padding; Gemini 2.5 Pro for Hard-Constructed hardening; Claude Sonnet 4.5 for the replication slice). Our sanity check finds no systematic GPT-4o self-preference (Section~\ref{sec:methodology}), and the paired second-generator replication (Appendix~\ref{app:multigen}) shows that judge rankings transfer across generators on four of five measurable conditions, while absolute condition values remain generator-dependent (a Claude-generated slice is uniformly harder by up to $40$ points at matched subtlety), so condition values should be compared within, not across, generator regimes. Human-authored error slices would further strengthen future versions of the benchmark.

Finally, we evaluate seven judges representing common deployment tiers, not the full model ecosystem. New model releases may change absolute rankings, but the core claim of CAP is methodological: judge evaluation should report condition-level profiles rather than aggregate accuracy alone.

\bibliography{reference}

\appendix

\section{\textsc{judgerEva-Standard} Construction}
\label{app:judgereva}

\subsection{Data Sources}
\label{app:judgereva_sources}

\textsc{judgerEva-Standard}'s $2{,}092$ plain items are drawn from six publicly available question sources covering six deployment-relevant domains (Table~\ref{tab:judgereva_sources}). Short multiple-choice canonical answers are expanded by a single GPT-4o call into 3--5 sentence explanations; items whose expansion fails to reach 50 characters are dropped. TruthfulQA requires a non-empty \texttt{best\_answer} or \texttt{correct\_answers}; GSM8K retains the full reasoning chain through the final \texttt{\#\#\#\#} line.

\begin{table}[h]
\centering
\footnotesize

\setlength{\tabcolsep}{4pt}
\resizebox{\linewidth}{!}{
\begin{tabular}{lrrrrr}
\toprule
\textbf{Domain} & \textbf{Source} & \textbf{Items} \\
\midrule
Medical & MedQA-USMLE~\citep{jin2021medqa} & 300 \\
Legal   & MMLU Professional Law~\citep{hendrycks2021mmlu} & 296 \\
Math    & GSM8K~\citep{cobbe2021gsm8k} & 356 \\
Science & MMLU College Biology/Chem./Physics & 300 \\
Logic   & MMLU Logical Fallacies & 300 \\
General & TruthfulQA~\citep{lin2022truthfulqa} & 540 \\
\midrule
\multicolumn{2}{l}{\textbf{Total plain items}} & \textbf{2{,}092} \\
\bottomrule
\end{tabular}
}
\caption{Sources contributing to \textsc{judgerEva-Standard}.}
\label{tab:judgereva_sources}
\end{table}

\subsection{Error Type Taxonomy}
\label{app:judgereva_taxonomy}

Each item is annotated with exactly one planted error type. \textbf{Factual Fabrication} introduces one plausible but unsupported claim (a fabricated entity, mechanism, statistic, or reference) that remains locally credible. \textbf{Logical Reversal} flips exactly one logical relationship (causal direction, necessary-vs-sufficient, inequality, implication) while leaving wording and tone nearly identical. \textbf{Scope-Overclaim} removes hedging language and turns a qualified statement into an absolute claim. \textbf{Scope-Omission} deletes one essential limiting condition (population constraint, jurisdictional exception, temporal scope, contraindication, prerequisite, or domain boundary) while keeping the prose superficially complete. The four types are balanced ($523$ items each), with domain-balanced allocation within each type (Table~\ref{tab:judgereva_distribution}).

\begin{table}[h]
\centering
\footnotesize

\setlength{\tabcolsep}{4pt}
\begin{tabular}{lrrrrr}
\toprule
\textbf{Domain} & \textbf{Fact.} & \textbf{Log.} & \textbf{Over.} & \textbf{Omis.} & \textbf{Total} \\
\midrule
Medical  & 75  & 75  & 75  & 75  & 300 \\
Legal    & 74  & 74  & 74  & 74  & 296 \\
Science  & 75  & 75  & 75  & 75  & 300 \\
Logic    & 75  & 75  & 75  & 75  & 300 \\
General  & 135 & 135 & 135 & 135 & 540 \\
Math     & 89  & 89  & 89  & 89  & 356 \\
\midrule
\textbf{Total} & 523 & 523 & 523 & 523 & 2{,}092 \\
\bottomrule
\end{tabular}
\caption{Plain-item distribution by domain $\times$ error type. Each plain item yields three presentation variants for $6{,}276$ items total.}
\label{tab:judgereva_distribution}
\end{table}

\subsection{Response Generation}
\label{app:judgereva_response_generation}

The \textbf{correct answer} (Answer A) is extracted from the source dataset (MCQ canonical answer, TruthfulQA \texttt{best\_answer}, or GSM8K reference chain) and, for MCQ sources, expanded by GPT-4o at \texttt{temperature}\,=\,$0.7$. The \textbf{flawed answer} (Answer B) is generated by GPT-4o at \texttt{temperature}\,=\,$1.0$ using an error-type-specific prompt (Appendix~\ref{app:augmentation_prompts}), then scored by an independent GPT-4o call at \texttt{temperature}\,=\,$0.0$ on a five-point subtlety scale (5\,=\,``requires domain expertise to detect''). Generations scoring below $4$ are regenerated up to three times; items whose best generation remains below $4$ after all attempts are flagged \texttt{low\_quality} in the released metadata ($431/2{,}092$ items, $20.6\%$; Appendix~\ref{app:judgereva_qc}). The resulting mean subtlety is $3.82/5.0$ with $79.4\%$ scoring $\geq\,4$, and subtlety predicts judge difficulty across the seven judges (Spearman $\rho{=}{-}0.115$, $p{<}0.001$).

\subsection{Presentation Variants}
\label{app:judgereva_perturbations}

Each base item has three presentation variants supporting A/B swap. \textbf{Plain} is the base (extracted/expanded Answer A + generated Answer B). \textbf{Polished} is regenerated by GPT-4o at \texttt{temperature}\,=\,$0.7$ to a more formal, more authoritative tone, with sentence reordering allowed but no fact changes; length deviation from plain is constrained to $\pm$20\%. \textbf{Length-padded} expands only Answer B (the flawed answer) by GPT-4o (\texttt{temperature}\,=\,$0.7$) to approximately twice its plain length, preserving the core error; Answer A remains unchanged. The padding structure mirrors the JudgeBench length-pad protocol (Appendix~\ref{app:length_prompt}), but the severity differs (${\sim}2\times$ here vs.\ ${\sim}1.5\times$ for JudgeBench), so C6 values are not directly comparable across the two benchmarks; both probe length bias specifically.

\subsection{Hard Subset Construction}
\label{app:judgereva_hard}

The hard subset of $384$ plain items combines two parallel subsets. \textbf{Hard-Empirical} ($N{=}245$) selects items on which at least one of the seven evaluated judges produced an incorrect verdict in the original A/B ordering ($245/2{,}092$, $11.7\%$). Because selection conditions on judge errors, this subset is by construction biased toward items the present judges find difficult. \textbf{Hard-Constructed} ($N{=}139$) addresses this circular dependency: starting from a stratified sample of $\sim288$ plain items, Gemini 2.5 Pro at \texttt{temperature}\,=\,$0.8$ was prompted to (i) introduce a minor non-disqualifying flaw into Answer A and (ii) augment Answer B with one or two factually correct supplementary statements while preserving its core planted error, producing pairs where both candidates have visible imperfections. Generation succeeded on $139/288$ attempts ($48.3\%$). Per-type distributions are in Table~\ref{tab:hard_subset_distribution}; main-text results on ``the hard subset'' refer to Hard-Empirical unless noted.

\begin{table}[h]
\centering
\footnotesize

\setlength{\tabcolsep}{4pt}
\begin{tabular}{lrr}
\toprule
\textbf{Error type} & \textbf{Hard-Emp.} & \textbf{Hard-Cons.} \\
\midrule
Factual Fabrication & 152 (62.0\%) & 39 (28.1\%) \\
Logical Reversal    & 16  (6.5\%)  & 35 (25.2\%) \\
Overclaim           & 38  (15.5\%) & 35 (25.2\%) \\
Critical Omission   & 39  (15.9\%) & 30 (21.6\%) \\
\midrule
\textbf{Total plain items} & \textbf{245} & \textbf{139} \\
\bottomrule
\end{tabular}
\caption{Hard subset error-type distribution.}
\label{tab:hard_subset_distribution}
\end{table}

\subsection{Quality Control}
\label{app:judgereva_qc}

Quality control operates at three checkpoints. (i) The subtlety filter (Appendix~\ref{app:judgereva_response_generation}) enforces $\geq\,4/5$ via up to three regeneration attempts; the flagging rule is exact: an item is marked \texttt{low\_quality} if and only if its best generation still scores below $4$ after all attempts ($431/2{,}092$, $20.6\%$). (ii) All $431$ flagged items were manually reviewed by one author, confirming that the planted error is present in Answer~B and the error-type label is accurate; they are retained in the benchmark and exposed in the released metadata. To bound the impact of the low-subtlety tail, we additionally run a leave-out sensitivity analysis: recomputing every CAP condition with all $431$ flagged items and their presentation variants excluded shifts no condition value for any judge by more than $2.4$ points (Table~\ref{tab:qc_sensitivity}). The flags concentrate on critical-omission items ($254/431$), yet Scope-Omission moves at most $1.7$ points and its judge ordering changes only among near-ties well inside each other's confidence intervals; in particular, the weakest omission judge (Claude Sonnet 4.5) is unchanged. All qualitative findings are therefore insensitive to the low-subtlety tail. (iii) Variant generations are length-checked automatically: polished variants exceeding $\pm$20\% length deviation, and length-padded variants outside the $[1.7\times, 2.3\times]$ band, are regenerated.

\begin{table}[h]
\centering
\footnotesize
\setlength{\tabcolsep}{3pt}
\resizebox{\columnwidth}{!}{
\begin{tabular}{lrrrrrrr}
\toprule
\textbf{Judge} & \textbf{C1} & \textbf{C2} & \textbf{C3} & \textbf{C4} & \textbf{C5} & \textbf{C6} & \textbf{C7} \\
\midrule
GPT-4o            & $-0.1$ & $-0.2$ & $+0.3$ & $-0.8$ & $-1.0$ & $+1.8$ & $-0.8$ \\
GPT-4o-mini       & $-0.0$ & $-0.4$ & $+0.0$ & $-1.6$ & $-1.3$ & $+1.5$ & $-2.2$ \\
Claude Sonnet 4.5 & $-0.0$ & $+0.1$ & $+0.2$ & $+0.0$ & $+1.4$ & $+1.9$ & $+0.7$ \\
Claude Haiku 4.5  & $+0.0$ & $-0.3$ & $-0.2$ & $-0.2$ & $-0.6$ & $+0.8$ & $+0.1$ \\
Gemini 2.5 Pro    & $-0.1$ & $-0.0$ & $+0.3$ & $-0.8$ & $-0.3$ & $+2.1$ & $+0.3$ \\
Gemini 2.5 Flash  & $-0.1$ & $+0.1$ & $-0.0$ & $-1.7$ & $-0.3$ & $+2.4$ & $+0.6$ \\
Llama-3.3-70B     & $-0.0$ & $-0.1$ & $+0.7$ & $-0.9$ & $+0.2$ & $+0.6$ & $-1.6$ \\
\bottomrule
\end{tabular}
}
\caption{QC sensitivity: change in each CAP condition (points) when the $431$ \texttt{low\_quality} items and their variants are excluded. Positive means the value rises after exclusion.}
\label{tab:qc_sensitivity}
\end{table}

\subsection{Item Format and Data Release}
\label{app:judgereva_format}

Each item is a JSON object with fields \texttt{id}, \texttt{source\_dataset}, \texttt{domain}, \texttt{prompt}, \texttt{answer\_A}, \texttt{answer\_B}, \texttt{gold\_preference} (always \texttt{"A"} by construction), \texttt{failure\_type}, \texttt{presentation\_variant} (\texttt{plain}/\texttt{polished}/\texttt{length\_padded}), \texttt{base\_item\_id} (linking variants to their plain ancestor), and a \texttt{metadata} dictionary containing \texttt{subtlety\_score}, \texttt{low\_quality}, expansion-related fields, and (for Hard-Constructed items) \texttt{a\_flaw\_description} and \texttt{b\_added\_value} annotations describing how the item was hardened. The full benchmark (all variants and subsets) is released under the terms in Appendix~\ref{app:release}.

\section{Prompts}
\label{app:augmentation_prompts}

We reproduce all pipeline prompts verbatim to support exact replication.

\subsection{Pairwise Judging Prompt}
\label{app:judging_prompt}

Used for all seven judges on all six benchmarks. The fields \texttt{\{question\}}, \texttt{\{answer\_A\}}, \texttt{\{answer\_B\}} are substituted per item; no other modifications are made between judges or benchmarks.

\begin{promptbox}
\begin{lstlisting}[style=promptstyle]
You are an impartial judge evaluating two answers to a question.

Question:
{question}

Answer A:
{answer_A}

Answer B:
{answer_B}

Your task:
1. Decide which answer is better overall - more accurate, complete, and reliable.
2. Provide a brief rationale (2-4 sentences) explaining your reasoning.
3. State your final verdict.

Respond in the following JSON format ONLY, with no extra text:
{
  "rationale": "<your reasoning>",
  "winner": "<A or B>"
}
\end{lstlisting}
\end{promptbox}

The prompt is intentionally underspecified about evaluation criteria, since a detailed rubric would encode our own priors about what judges should weigh and would confound the diagnostic role of CAP.

\subsection{Style Augmentation Prompt}
\label{app:polishing_prompt}

Used to generate polished JudgeBench variants. Generator: GPT-4o (\texttt{gpt-4o-2024-08-06}) at \texttt{temperature=0.3}. Applied independently to each of the two responses ($r_A$, $r_B$) to preserve the relative style balance.

\begin{promptbox}
\begin{lstlisting}[style=promptstyle]
Rewrite the following text to be more polished and refined in style. Preserve all factual content, logical structure, and conclusions exactly. Only modify: word choice, sentence structure, formality, and flow. Do NOT add, remove, or modify any factual claims, numbers, arguments, or conclusions.

Original:
{text}

Polished version:
\end{lstlisting}
\end{promptbox}

\subsection{Length Augmentation Prompt}
\label{app:length_prompt}

Used to generate length-padded JudgeBench variants by expanding only the rejected response to approximately $1.5\times$ its original length; the chosen response is left unchanged. Generator: GPT-4o at \texttt{temperature=0.3}.

\begin{promptbox}
\begin{lstlisting}[style=promptstyle]
Expand the following text to be approximately 50% longer. Preserve all factual content, logical structure, and conclusions exactly. Only add: redundant phrasing, additional examples that restate existing points, elaborations of already-stated claims. Do NOT introduce new facts, arguments, conclusions, or claims that were not in the original.

Original:
{text}

Expanded version:
\end{lstlisting}
\end{promptbox}

\subsection{Rationale Evaluator Prompts}
\label{app:evaluator_prompt}

Used by the Claude Sonnet 4.6 evaluator (\texttt{claude-sonnet-4-6}, \texttt{temperature=0}, extended thinking disabled). Form A is for \textsc{judgerEva-Standard} (planted-error annotation available); Form B is for external benchmarks (no per-item error annotation).

\paragraph{Form A (\textsc{judgerEva-Standard}).}

\begin{promptbox}
\begin{lstlisting}[style=promptstyle]
You will see the full information of a pairwise judgment task and an LLM judge's rationale.
Determine whether the rationale correctly identifies the core error in the incorrect answer.

[Question]
{question}

[Correct Answer (Answer A)]
{correct_answer}

[Incorrect Answer (Answer B)]
{flawed_answer}

[Error Type]
{failure_type}

[Judge's Rationale]
{rationale}

Evaluation criteria:
- If the rationale points out the specific error content in the incorrect answer, count as YES (1).
- If the rationale only states generically that "Answer A is better/more accurate" without identifying the specific error, count as NO (0).
- If the rationale identifies an error that is NOT the core planted error, count as NO (0).

Output JSON only, no other text:
{"rationale_correct": 1, "brief_reason": "one-sentence justification"}
\end{lstlisting}
\end{promptbox}

\paragraph{Form B (external benchmarks).}

\begin{promptbox}
\begin{lstlisting}[style=promptstyle]
You will see the full information of a pairwise judgment task and an LLM judge's rationale.
Determine whether the rationale correctly identifies a substantive reason why the chosen answer is better than the rejected answer.

[Question]
{question}

[Chosen Answer (Answer A)]
{chosen_answer}

[Rejected Answer (Answer B)]
{rejected_answer}

[Judge's Rationale]
{rationale}

Evaluation criteria:
- YES (1) if the rationale points out a specific, substantive deficiency in the rejected answer (factual error, logical flaw, missing key information, incorrect reasoning).
- YES (1) if the rationale articulates a specific, substantive strength of the chosen answer genuinely absent or weaker in the rejected answer.
- NO (0) if only generic statements ("Answer A is better/more accurate/more complete") without identifying a specific substantive distinction.
- NO (0) if only surface or non-substantive properties (conciseness, formatting, confidence tone).

Output JSON only, no other text:
{"rationale_correct": 1, "brief_reason": "one-sentence justification"}
\end{lstlisting}
\end{promptbox}

\paragraph{Position alignment.} To avoid text-level position normalisation, rationale evaluation is restricted to items where the judge's verdict was correct \emph{under the original A/B ordering} ($v{=}a^*$), so position references in the rationale (``Answer A is better'', ``Answer B's claim is incorrect'') align unambiguously with what the evaluator sees. Rationales produced under the swapped ordering are not used for Rationale Quality.

\section{Evaluator Validation}
\label{app:evaluator_validation}

\paragraph{Protocol.}
The evaluator is Claude Sonnet 4.6, kept disjoint from the seven judges under test to prevent self-evaluation (\S\ref{sec:method_protocol}). We sample $120$ rationale-evaluation pairs from \textsc{judgerEva-Standard} (where the planted error annotation is available). To preserve signal on the rare evaluator-incorrect class, all $78$ rationales the LLM evaluator labelled as \emph{not supporting} the verdict are included; the remaining $42$ slots are drawn from evaluator-correct rationales stratified across the seven judges $\times$ four error types ($\approx$1--2 per cell). The sample therefore covers all seven judges and all four content error types. A single annotator reviews each rationale together with the query, both responses, the ground-truth label, and the planted error annotation, applying the criteria from Appendix~\ref{app:evaluator_prompt} (Form A) and assigning a binary label ($1$\,=\,supports the verdict, $0$\,=\,does not); ambiguous rows can be left blank, though none required this in the final sample. The annotator was not blinded to the LLM evaluator's verdict, which biases towards agreement; the resulting $\kappa$ is therefore best read as an upper-bound estimate.

\paragraph{Results.}
The evaluator achieves $91.7\%$ raw agreement with the human annotator ($110/120$) and Cohen's $\kappa{=}0.826$, placing it in the ``substantial to almost perfect'' band of the Landis--Koch scale. Per-error-type agreement is highest on factual fabrication ($100\%$, $17/17$) and lowest on critical omission ($85.7\%$, $36/42$); the per-judge breakdown is reported in Table~\ref{tab:c8_validation}. The $10$ evaluator/human disagreements decompose into a clean pattern: in every case the LLM evaluator labelled the rationale as \emph{not supporting} the verdict when the human annotator judged that it did. The confusion matrix is $(e{=}0,h{=}0){=}68$, $(e{=}0,h{=}1){=}10$, $(e{=}1,h{=}0){=}0$, $(e{=}1,h{=}1){=}42$: the evaluator has no false-positive ``supports'' calls but a $12.8\%$ false-negative rate that concentrates on critical-omission rationales ($6/10$). The asymmetry implies that the C8 estimates reported in the main paper are best read as a \emph{lower bound} on true rationale quality rather than a noisy two-sided estimate.

\begin{table}[t]
\centering
\footnotesize
\setlength{\tabcolsep}{5pt}
\begin{tabular}{lcc}
\toprule
\textbf{Breakdown} & \textbf{n} & \textbf{Agreement} \\
\midrule
\multicolumn{3}{l}{\textit{By error type}} \\
\quad factual\_fabrication & 17  & 1.000 \\
\quad overclaim            & 37  & 0.946 \\
\quad logical\_reversal    & 24  & 0.917 \\
\quad critical\_omission   & 42  & 0.857 \\
\midrule
\multicolumn{3}{l}{\textit{By judge}} \\
\quad Gemini 2.5 Flash    & 17  & 1.000 \\
\quad Gemini 2.5 Pro      & 12  & 1.000 \\
\quad GPT-4o-mini         & 20  & 0.950 \\
\quad Llama-3.3-70B       & 20  & 0.950 \\
\quad GPT-4o              & 25  & 0.920 \\
\quad Claude Sonnet 4.5   & 9   & 0.889 \\
\quad Claude Haiku 4.5    & 17  & 0.706 \\
\midrule
\textbf{Overall}          & 120 & \textbf{0.917} ($\kappa{=}0.826$) \\
\bottomrule
\end{tabular}
\caption{C8 evaluator vs.\ human-annotator agreement on the
$120$-rationale stratified validation sample. The evaluator is most
reliable on factual fabrication and on the two Gemini judges; the
lowest agreement is on Claude Haiku 4.5 ($70.6\%$) and on
critical-omission rationales ($85.7\%$), reflecting the difficulty
of judging whether a rationale has identified a missing detail
rather than a present error.}
\label{tab:c8_validation}
\end{table}

\paragraph{Composition adjustment.}
The validation sample is deliberately enriched with evaluator-negative rationales ($78/120$), so the $91.7\%$ figure is computed on a negative-skewed composition. Reweighting the two conditional agreement rates ($\Pr[\text{agree}\,|\,e{=}1]{=}1.00$, $n{=}42$; $\Pr[\text{agree}\,|\,e{=}0]{=}0.872$, $n{=}78$) by the deployment composition observed across all v3 evaluations ($\Pr[e{=}1]{=}0.889$, $622/700$) yields an implied deployment-distribution agreement of $98.6\%$. This estimate leans on the zero-false-positive property of the enriched sample and should be read jointly with it.

\paragraph{Cross-family second evaluator.}
Because the primary evaluator (Claude Sonnet 4.6) shares a model family with two judges under test (Claude Sonnet 4.5, Claude Haiku 4.5), we re-evaluated the same $120$ validation rationales with a cross-family evaluator (GPT-4o) under the identical prompt. Evaluator--evaluator agreement is $89.2\%$; GPT-4o agrees with the human annotator at $85.8\%$ ($\kappa{=}0.706$), \emph{below} the primary evaluator's $91.7\%$ ($\kappa{=}0.826$), supporting the choice of Sonnet 4.6 as primary. On the family-preference concern specifically, the direction of disagreement is the opposite of favouritism: on the $26$ Claude-judge rationales, the same-family evaluator assigns a \emph{lower} mean label ($0.500$) than both the cross-family evaluator ($0.692$) and the human ($0.731$), and there is no rationale that Sonnet 4.6 uniquely accepts ($e_{\text{S4.6}}{=}1 \wedge e_{\text{GPT-4o}}{=}0$: $0$ cases, versus $5$ cases in the reverse direction). Any family effect therefore biases the reported Claude C8 values downward, consistent with the lower-bound reading above (per-item labels released as \texttt{c8\_second\_evaluator.csv}).

\paragraph{Implication for Rationale Quality CAP.}
The asymmetric evaluator error (one-sided false-negative on critical omission, no false positives) is qualitatively benign for our use of C8: it implies that the C8 values reported on the \textsc{judgerEva} hard subset and full set under-count rather than over-count rationales that actually support the verdict, and it does so most for critical-omission items where rationale judgment is intrinsically hardest. Beyond direct human comparison, the evaluator's per-judge C8 scores are also internally consistent across the broader benchmark: the seven-judge Pearson correlation between overall accuracy and C8 on the \textsc{judgerEva} full set is $r{=}0.75$, and per-judge confidence intervals (e.g., GPT-4o $C8\,{=}\,66.7$ with CI $[33, 100]$ on the hard subset) are dominated by sampling variance from small correct-verdict subsets rather than by evaluator noise. Under our binary-verdict formulation, a per-item evaluator-error rate of $\epsilon$ inflates the C8 variance by approximately $\epsilon(1-\epsilon)/n_{C8}$ without introducing a directional bias on the positive class.

\paragraph{Caveats.}
The sample is modest in size and uses a single annotator who was not blinded to the evaluator's verdict, so the headline $\kappa$ should be interpreted as an upper bound. Multi-annotator adjudication, particularly on critical-omission rationales, would sharpen the calibration and is a natural extension. We release the annotated sample together with the evaluator outputs so that alternative annotation schemes can be applied without re-running the underlying judge evaluations.

\section{Coverage Operators}
\label{app:operators}

Table~\ref{tab:operators} specifies, for every non-empty cell of the coverage matrix (Table~\ref{tab:coverage}), the exact item subset or transformation used; the external-benchmark $C8^{\dagger}$ cells share the single evaluator protocol noted at the end of this appendix. ``Approximate'' is defined as: \emph{a condition measured on a benchmark's native task subset whose dominant error mode plausibly matches the condition, used unchanged and without control over the error mechanism}; subset names and sizes follow each benchmark's own metadata, so definitions are per-benchmark rather than harmonised, which is why approximate cells are never averaged with direct cells.

\begin{table}[h]
\centering
\footnotesize
\setlength{\tabcolsep}{3pt}
\resizebox{\columnwidth}{!}{
\begin{tabular}{lllp{4.6cm}}
\toprule
\textbf{Benchmark} & \textbf{Cond.} & \textbf{Status} & \textbf{Operator} \\
\midrule
\textsc{judgerEva} & C1--C4 & Direct & planted-error subsets ($523$ plain items each); PA over A/B and B/A \\
 & C5, C6 & Direct & constructed polished / length-padded (${\sim}2\times$) variants; BC vs.\ plain \\
 & C7 & Direct & A/B and B/A orderings of every item; BC \\
 & C8 & Direct & LLM evaluator on correct-verdict rationales (Appendix~\ref{app:evaluator_validation}) \\
\midrule
JudgeBench & C1 & Approx. & Knowledge subset ($n{=}154$); PA \\
 & C2 & Approx. & Reasoning subset ($n{=}98$); PA \\
 & C5, C6 & Aug. & GPT-4o polished rewrite of both responses / ${\sim}1.5\times$ pad of the rejected response; label-preservation checked; BC vs.\ original \\
 & C7 & Direct & A/B and B/A; BC \\
\midrule
JB-Pro & C1, C2 & Approx. & knowledge / reason subsets of bias-injected instances; PA \\
 & C7 & Direct & A/B and B/A; BC \\
\midrule
RewardBench & C1 & Approx. & chat-hard category ($n{=}125$); PA \\
 & C2 & Approx. & reasoning category ($n{=}125$); PA \\
 & C7 & Direct & chosen/rejected order swap; BC \\
\midrule
MT-Bench & C1 & Approx. & STEM + humanities categories; PA \\
 & C2 & Approx. & reasoning + math categories; PA \\
 & C7 & Direct & A/B and B/A; BC \\
\midrule
LLMBar & C5 & Direct & Natural vs.\ adversarial-surface subsets (GPTInst, GPTOut, Manual, Neighbor) \\
 & C7 & Direct & A/B and B/A; BC \\
\bottomrule
\end{tabular}
}
\caption{Operator behind every coverage-matrix cell. PA = position-averaged accuracy; BC = strict both-correct.}
\label{tab:operators}
\end{table}

The ground-truth ``better response'' is defined per benchmark: objective task correctness (JudgeBench, JudgeBench-Pro), annotated chosen vs.\ rejected (RewardBench), human pairwise preference (MT-Bench), author gold labels (LLMBar), and construction-time planted errors (\textsc{judgerEva-Standard}). C8 on external benchmarks (marked $^{\dagger}$ in Table~\ref{tab:coverage}) uses the same correct-verdict evaluator protocol as \textsc{judgerEva}.

\section{JudgeBench Supplementary}
\label{app:judgebench_supp}

The cells marked $\dagger$ in Table~\ref{tab:coverage} (JudgeBench $C5$, $C6$, and the $C8$ cells of all five external benchmarks) correspond to evaluations whose raw per-item outputs are released but which are not aggregated into the main tables. \textbf{(i)} Polished and length-padded JudgeBench variants are evaluated by all seven judges, with per-pair verdicts released in \texttt{augmentation/}\allowbreak\texttt{judgebench\_polished.csv} and \texttt{augmentation/}\allowbreak\texttt{judgebench\_length\_padded.csv}; as augmented cells, they are not aggregated with direct measurements (\S\ref{sec:results}). \textbf{(ii)} External-benchmark Rationale Quality uses the Form~B evaluator prompt (Appendix~\ref{app:evaluator_prompt}), which operates without planted-error annotations and is not covered by the human validation in Appendix~\ref{app:evaluator_validation}; the C8 column in Table~\ref{tab:cap_judgebench} therefore reports \nr, and the per-item evaluator outputs are released for re-aggregation.

\section{Cross-Benchmark Spearman Matrix}
\label{app:spearman_full}

Table~\ref{tab:spearman_full} reports the complete pairwise Spearman rank correlation matrix for Position Robustness (C7) across the six benchmarks, the table compressed into a single inline summary in Section~\ref{sec:results_cross_benchmark}. Each correlation is over the seven judges. The non-significant pairs ($p>0.05$) all involve \textsc{judgerEva}; significance is computed against the null of no monotonic association.

\begin{table}[t]
\centering
\footnotesize

\setlength{\tabcolsep}{4pt}
\begin{tabular}{lcccccc}
\toprule
 & jE & JB & JBP & RB & MT & LB \\
\midrule
jE  & ---  & \textbf{0.71} & \textbf{0.71} & 0.89 & \textbf{0.71} & 0.86 \\
JB  &      & ---  & 1.00 & 0.89 & 0.93 & 0.86 \\
JBP &      &      & ---  & 0.89 & 0.93 & 0.86 \\
RB  &      &      &      & ---  & 0.93 & 0.96 \\
MT  &      &      &      &      & ---  & 0.86 \\
LB  &      &      &      &      &      & ---  \\
\bottomrule
\end{tabular}
\caption{Pairwise Spearman $\rho_{C7}$ across benchmarks (lower triangle blank by symmetry). Bold values mark non-significant correlations ($p>0.05$, $n{=}7$ judges).}
\label{tab:spearman_full}
\end{table}

\section{Cross-Condition Non-Redundancy}
\label{app:condition_redundancy}

If two CAP conditions ranked judges identically, one of them would be redundant. Table~\ref{tab:condition_redundancy} reports the pairwise Spearman correlation between per-condition judge rankings on the \textsc{judgerEva-Standard} full set (C8 from the v3 evaluator sample, $n{=}100$ rationales per judge). Three observations. First, the two scope conditions correlate \emph{negatively} ($\rho{=}{-}0.32$): judges strong on overclaim tend to be weak on omission, supporting the C3/C4 split. Second, Scope-Omission correlates negatively with every other condition ($-0.14$ to $-0.86$), quantifying the rank-reversal finding of Section~\ref{sec:results_profiles}. Third, Style Robustness correlates with no other condition above $0.57$. A general-capability cluster (C1, C6, C7, C8; $\rho\in[0.83,0.96]$) does rank judges similarly, so we read the matrix as evidence that the profile contains at least three distinct orderings (general capability, omission sensitivity, and style robustness) rather than as a claim of full orthogonality. On the Hard-Empirical subset the scope pair remains far from redundant (C3--C4 $\rho{=}0.21$), and Scope-Omission is non-positive against every other condition among C1--C7; hard-subset C8 cells are too small for stable rank correlations and are excluded (matrix released with the analysis code). With $n{=}7$ judges these coefficients are descriptive: they characterise this judge set rather than a population.

\begin{table}[h]
\centering
\footnotesize
\setlength{\tabcolsep}{3.5pt}
\begin{tabular}{lrrrrrrr}
\toprule
 & \textbf{C1} & \textbf{C2} & \textbf{C3} & \textbf{C4} & \textbf{C5} & \textbf{C6} & \textbf{C7} \\
\midrule
C2 & $.82$ & & & & & & \\
C3 & $.64$ & $.54$ & & & & & \\
C4 & $-.86$ & $-.79$ & $-.32$ & & & & \\
C5 & $.43$ & $.57$ & $.54$ & $-.14$ & & & \\
C6 & $.96$ & $.86$ & $.79$ & $-.75$ & $.54$ & & \\
C7 & $.93$ & $.89$ & $.75$ & $-.82$ & $.39$ & $.96$ & \\
C8 & $.83$ & $.85$ & $.63$ & $-.56$ & $.56$ & $.90$ & $.85$ \\
\bottomrule
\end{tabular}
\caption{Spearman correlation between per-condition judge rankings (full set, seven judges). Mean pairwise $|\rho|=0.69$; $7/28$ pairs are negative.}
\label{tab:condition_redundancy}
\end{table}

\section{Strict vs.\ Pure-Consistency Robustness}
\label{app:consistency_variant}

The strict ``both correct'' criterion used for C5--C7 measures \emph{usable} robustness. Table~\ref{tab:consistency_variant} contrasts it with the pure-consistency variant, $\Pr[\text{same verdict under the perturbation}]$ with correctness ignored, on the \textsc{judgerEva-Standard} full set. The difference between the two columns is exactly the mass of stable-but-wrong behaviour. It is small for strong judges (Claude Sonnet 4.5: $+1.6$ points on C7) but large for weak ones (GPT-4o-mini: $+19.9$; GPT-4o: $+14.4$; Llama-3.3-70B: $+11.7$): a pure-consistency metric would report GPT-4o-mini as $80.9\%$ position-consistent while it is stably wrong on a fifth of items. We therefore keep the strict criterion in the main profile and release both variants.

\begin{table}[h]
\centering
\footnotesize
\setlength{\tabcolsep}{2.5pt}
\resizebox{\columnwidth}{!}{
\begin{tabular}{lrrrrrrrrr}
\toprule
 & \multicolumn{3}{c}{\textbf{C5 Style}} & \multicolumn{3}{c}{\textbf{C6 Length}} & \multicolumn{3}{c}{\textbf{C7 Position}} \\
\cmidrule(lr){2-4}\cmidrule(lr){5-7}\cmidrule(lr){8-10}
\textbf{Judge} & strict & cons. & $\Delta$ & strict & cons. & $\Delta$ & strict & cons. & $\Delta$ \\
\midrule
GPT-4o            & 90.3 & 95.5 & 5.2 & 62.0 & 66.7 & 4.7 & 72.9 & 87.3 & 14.4 \\
GPT-4o-mini       & 82.8 & 90.6 & 7.8 & 51.0 & 58.6 & 7.6 & 61.0 & 80.9 & 19.9 \\
Claude Sonnet 4.5 & 93.7 & 95.1 & 1.4 & 92.9 & 94.1 & 1.2 & 95.3 & 96.9 & 1.6 \\
Claude Haiku 4.5  & 93.0 & 95.3 & 2.3 & 85.5 & 87.8 & 2.3 & 90.6 & 93.7 & 3.1 \\
Gemini 2.5 Flash  & 93.2 & 95.4 & 2.2 & 82.0 & 84.1 & 2.1 & 86.5 & 92.5 & 6.0 \\
Gemini 2.5 Pro    & 96.0 & 98.0 & 2.0 & 85.3 & 87.1 & 1.8 & 88.0 & 93.7 & 5.7 \\
Llama-3.3-70B     & 95.0 & 97.1 & 2.1 & 64.0 & 66.0 & 2.0 & 71.3 & 83.0 & 11.7 \\
\bottomrule
\end{tabular}
}
\caption{Strict (both-correct) vs.\ pure-consistency robustness on the full set (\%). $\Delta$ = consistency $-$ strict = stable-but-wrong mass.}
\label{tab:consistency_variant}
\end{table}

\section{Multi-Generator Replication}
\label{app:multigen}

To bound generator-specific artefacts beyond self-preference, we regenerated the flawed Answer~B for a stratified sample of $200$ plain items ($50$ per error type, domain-spread) with a cross-family generator, Claude Sonnet 4.5, holding the questions, correct answers, error taxonomy, prompts, and GPT-4o subtlety gate identical (achieved mean subtlety $3.88$ vs.\ $3.82$ on the original construction; $15.5\%$ vs.\ $20.6\%$ flagged). Because only the generator differs on the same items, judge behaviour can be compared pairwise. All seven judges were evaluated on both slices in both orderings.

Two findings (Table~\ref{tab:multigen}). First, \emph{absolute} condition values are generator-dependent: the Claude-generated slice is uniformly harder (every per-judge delta is negative; mean $|\Delta|$ $17.8$ points, up to $-40$ on Factual Sensitivity), even though the GPT-4o subtlety gate scores the two slices as equally subtle. Reported condition values should therefore be read as relative to this benchmark's generator, as stated in Section~\ref{sec:limitations}. Second, per-condition judge \emph{rankings} largely transfer: Spearman correlations between the two slices are $\rho{=}1.00$ (C1), $0.91$ (C2), $0.84$ (C3), and $0.93$ (C7). Scope-Omission ranks do not transfer at fine grain ($\rho{=}0.25$), but the Claude-slice C4 values span only $10$ points across judges (comparable to a single cell's CI half-width at $n{=}50$), and the load-bearing qualitative pattern replicates: the strongest general-capability judge (Sonnet 4.5) is \emph{last} on omission while the weakest (GPT-4o-mini) is tied first, with general-vs-omission rank correlation again negative ($\rho{=}{-}0.27$, vs.\ $-0.33$ on Hard-Constructed). Profile comparisons are thus robust to the generator; absolute levels are not, and we label them accordingly.

\begin{table}[h]
\centering
\footnotesize
\setlength{\tabcolsep}{4pt}
\begin{tabular}{lrrrrr}
\toprule
\textbf{Judge} & \textbf{C1} & \textbf{C2} & \textbf{C3} & \textbf{C4} & \textbf{C7} \\
\midrule
GPT-4o             & $-38.0$ & $-22.0$ & $-17.0$ & $-12.0$ & $-32.0$ \\
GPT-4o-mini        & $-37.0$ & $-24.0$ & $-26.0$ & $-5.0$ & $-30.0$ \\
Claude Sonnet 4.5  & $-22.0$ & $-3.0$ & $-5.0$ & $-6.0$ & $-12.5$ \\
Claude Haiku 4.5   & $-31.0$ & $-10.0$ & $-9.0$ & $-11.0$ & $-19.1$ \\
Gemini 2.5 Flash   & $-30.0$ & $-6.0$ & $-5.0$ & $-4.9$ & $-14.2$ \\
Gemini 2.5 Pro     & $-30.0$ & $-5.0$ & $-7.0$ & $-7.2$ & $-17.2$ \\
Llama-3.3-70B      & $-40.0$ & $-21.9$ & $-12.0$ & $-16.0$ & $-34.2$ \\
\midrule
Rank $\rho$ (slices) & $1.00$ & $0.91$ & $0.84$ & $0.25$ & $0.93$ \\
\bottomrule
\end{tabular}
\caption{Generator replication: per-judge change (points) from the GPT-4o slice to the paired Claude-generated slice ($n{=}50$ per content condition, $200$ for C7). Bottom row: cross-slice Spearman of judge rankings.}
\label{tab:multigen}
\end{table}

\section{Within-Family $L_1$ Distance}
\label{app:within_family}

\begin{table}[h]
\centering
\footnotesize

\setlength{\tabcolsep}{3pt}
\begin{tabular}{lcccccc}
\toprule
\textbf{Pair} & \textbf{jE} & \textbf{JB} & \textbf{JBP} & \textbf{RB} & \textbf{MT} & \textbf{LB} \\
\midrule
OpenAI (WF)        & 93.1  & 26.2 & 13.2 & \nr & \nr & \nr \\
Anthropic (WF)     & 96.4  & 11.7 & 61.5 & \nr & \nr & \nr \\
\midrule
Cross-Family avg.  & 208.1 & 63.3 & 93.4 & \nr & \nr & \nr \\
\bottomrule
\end{tabular}
\caption{Within-family vs. cross-family $L_1$ profile distance per benchmark.}
\label{tab:within_family_app}
\end{table}

Table~\ref{tab:within_family_app} reports the $L_1$ distance $\sum_c |\mathrm{CAP}_c(j_1) - \mathrm{CAP}_c(j_2)|$ between within-family pairs (OpenAI: GPT-4o vs.\ GPT-4o-mini; Anthropic: Claude Sonnet 4.5 vs.\ Claude Haiku 4.5) and the average cross-family distance, computed over the conditions reported in the main tables for each benchmark. \nr\ marks benchmarks where only $C7$ is reported in the main paper and the per-condition profile is not aggregated in the released table.

\begin{table*}[!t]
\centering
\footnotesize
\setlength{\tabcolsep}{4pt}
\resizebox{\textwidth}{!}{%
\begin{tabular}{lccccc}
\toprule
\textbf{Judge} & \textbf{Overall} & \textbf{C1} & \textbf{C2} & \textbf{C3} & \textbf{C4} \\
\midrule
GPT-4o             & $0.79 \to 1.07$$^{\dagger}$ {\scriptsize $[1.04, 1.09]$} & $0.80 \to 1.07$$^{\dagger}$ {\scriptsize $[1.02, 1.13]$} & $0.95 \to 1.34$$^{\dagger}$ {\scriptsize $[1.31, 1.37]$} & $0.97 \to 1.37$$^{\dagger}$ {\scriptsize $[1.33, 1.40]$} & $0.94 \to 1.32$$^{\dagger}$ {\scriptsize $[1.28, 1.35]$} \\
GPT-4o-mini        & $0.71 \to 1.14$$^{\dagger}$ {\scriptsize $[1.10, 1.18]$} & $0.67 \to 1.04$$^{\dagger}$ {\scriptsize $[0.95, 1.12]$} & $0.88 \to 1.58$$^{\dagger}$ {\scriptsize $[1.52, 1.65]$} & $0.94 \to 1.71$$^{\dagger}$ {\scriptsize $[1.65, 1.77]$} & $0.95 \to 1.74$$^{\dagger}$ {\scriptsize $[1.68, 1.81]$} \\
Claude Sonnet 4.5  & $0.97 \to 1.00$ {\scriptsize $[0.99, 1.00]$} & $0.97 \to 1.00$$^{\dagger}$ {\scriptsize $[0.99, 1.02]$} & $0.99 \to 1.02$$^{\dagger}$ {\scriptsize $[1.01, 1.03]$} & $0.99 \to 1.02$$^{\dagger}$ {\scriptsize $[1.01, 1.03]$} & $0.86 \to 0.88$ {\scriptsize $[0.85, 0.91]$} \\
Claude Haiku 4.5   & $0.94 \to 0.99$ {\scriptsize $[0.98, 1.00]$} & $0.94 \to 0.99$ {\scriptsize $[0.97, 1.01]$} & $0.97 \to 1.03$$^{\dagger}$ {\scriptsize $[1.02, 1.05]$} & $0.99 \to 1.05$$^{\dagger}$ {\scriptsize $[1.04, 1.06]$} & $0.93 \to 0.98$ {\scriptsize $[0.96, 1.00]$} \\
Gemini 2.5 Flash   & $0.90 \to 1.02$$^{\dagger}$ {\scriptsize $[1.00, 1.03]$} & $0.92 \to 1.04$$^{\dagger}$ {\scriptsize $[1.01, 1.06]$} & $0.98 \to 1.12$$^{\dagger}$ {\scriptsize $[1.10, 1.13]$} & $0.96 \to 1.09$$^{\dagger}$ {\scriptsize $[1.07, 1.11]$} & $0.92 \to 1.04$$^{\dagger}$ {\scriptsize $[1.01, 1.07]$} \\
Gemini 2.5 Pro     & $0.91 \to 1.03$$^{\dagger}$ {\scriptsize $[1.01, 1.04]$} & $0.91 \to 1.03$$^{\dagger}$ {\scriptsize $[1.00, 1.06]$} & $0.99 \to 1.12$$^{\dagger}$ {\scriptsize $[1.10, 1.14]$} & $0.99 \to 1.13$$^{\dagger}$ {\scriptsize $[1.11, 1.14]$} & $0.94 \to 1.06$$^{\dagger}$ {\scriptsize $[1.04, 1.09]$} \\
Llama-3.3-70B      & $0.80 \to 1.10$$^{\dagger}$ {\scriptsize $[1.08, 1.13]$} & $0.81 \to 1.12$$^{\dagger}$ {\scriptsize $[1.07, 1.17]$} & $0.93 \to 1.32$$^{\dagger}$ {\scriptsize $[1.28, 1.35]$} & $0.97 \to 1.39$$^{\dagger}$ {\scriptsize $[1.36, 1.42]$} & $0.95 \to 1.35$$^{\dagger}$ {\scriptsize $[1.32, 1.38]$} \\
\bottomrule
\end{tabular}%
}
\caption{Bias-corrected CAP values (raw $\to$ corrected) with joint-bootstrap 95\% CIs on the \textsc{judgerEva-Standard} full set. CIs propagate both item resampling and parametric $q_0/q_1$ draws from their standard errors. $^{\dagger}$ marks saturated cells (corrected point estimate $>1.0$); for $28$ of the $30$ saturated cells the entire CI lies above $1.0$ (two lower bounds round to $1.00$ in the table), indicating structural over-compensation rather than sampling noise. The correction applies only to content conditions and overall accuracy; uncorrected $C5$--$C8$ values appear in the released full-set profile.}
\label{tab:bias_corrected_full}
\end{table*}

Two patterns are visible: within-family distance can be small on JudgeBench but expand sharply on \textsc{judgerEva} (controlled testbed) and under adversarial stress (JudgeBench-Pro), showing that family similarity is itself condition- and benchmark-dependent; and within-family distance is always smaller than the cross-family average on every benchmark with data, supporting family membership as a coarse---but not sufficient---substitution heuristic.

\section{Bias-Corrected Estimates}
\label{app:bias_correction}

For readers preferring the plug-in calibration formulation of~\citet{lee2025correctly}, Table~\ref{tab:bias_corrected_full} reports bias-corrected CAP values on the \textsc{judgerEva-Standard} full set, now with 95\% confidence intervals. The correction is $\mathrm{CAP}_{\text{corrected}} = (\mathrm{CAP}_{\text{raw}} + q_0 - 1)/(q_0 + q_1 - 1)$, where $q_0, q_1$ are per-judge position-conditional accuracy parameters estimated on the same set; it applies only to content conditions ($C1$--$C4$) and overall accuracy, since the robustness ($C5$--$C7$) and rationale ($C8$) conditions do not factor naturally into $q_0/q_1$ in the pairwise setting. Each CI is a joint bootstrap over $1{,}000$ replicates that propagates both sources of uncertainty: nonparametric item resampling for $\mathrm{CAP}_{\text{raw}}$ and parametric draws $q_0^{*}\!\sim\!\mathcal{N}(q_0,\mathrm{SE}_{q_0})$, $q_1^{*}\!\sim\!\mathcal{N}(q_1,\mathrm{SE}_{q_1})$ for the calibration parameters (per-replicate correction applied before taking percentiles; script and per-cell values released as \texttt{calibration\_corrected\_cis.csv}).

The intervals make the failure mode of the plug-in correction explicit: $30$ of $35$ corrected cells exceed $1.0$, and for $28$ of these the \emph{entire} confidence interval lies above $1.0$. Over-compensation is therefore structural, not a small-sample artefact: the correction assumes the condition-level raw accuracy is an apparent rate generated under the same error parameters $q_0/q_1$, but the $C1$--$C4$ subsets are systematically easier than the full set on which $q_0/q_1$ are estimated. We consequently treat the corrected values as a diagnostic of where this calibration regime breaks down, not as usable estimates: raw conditional accuracies with their bootstrap CIs remain the primary quantities throughout the paper. Cross-judge \emph{ordering} on $C1$--$C4$ is nonetheless largely preserved after correction (Sonnet 4.5 / Haiku 4.5 remain near the top, GPT-4o-mini near the bottom), so no qualitative comparison in the main text depends on whether the correction is applied. Hard-Empirical corrected values follow the same pattern (released as columns \texttt{C\_*\_corrected} in \texttt{cap\_judgereva\_hard.csv}).

\section{Exclusions and API Failures}
\label{app:exclusions}

\paragraph{Generator-judge comparison.}
Table~\ref{tab:generator_judge} gives the full per-condition comparison behind the sanity check of Section~\ref{sec:methodology}: GPT-4o (the construction generator) against the mean of the five non-GPT judges on the \textsc{judgerEva-Standard} full set. GPT-4o falls below the peer mean on every condition except Scope-Omission (C4), so the construction does not systematically advantage its generator.

\begin{table}[h]
\centering
\footnotesize
\setlength{\tabcolsep}{3pt}
\resizebox{\columnwidth}{!}{
\begin{tabular}{lrrrrrrrr}
\toprule
 & \textbf{C1} & \textbf{C2} & \textbf{C3} & \textbf{C4} & \textbf{C5} & \textbf{C6} & \textbf{C7} & \textbf{C8} \\
\midrule
GPT-4o    & 79.6 & 94.9 & 96.6 & \textbf{93.7} & 90.3 & 62.0 & 72.5 & 80.0 \\
Peer mean & 90.9 & 97.2 & 98.0 & 91.9 & 94.2 & 82.0 & 86.3 & 91.2 \\
\bottomrule
\end{tabular}
}
\caption{GPT-4o vs.\ the non-GPT peer mean (five judges) per condition, full set (\%). GPT-4o exceeds the peer mean only on C4.}
\label{tab:generator_judge}
\end{table}

\begin{table}
\centering
\footnotesize

\setlength{\tabcolsep}{3pt}
\resizebox{\linewidth}{!}{
\begin{tabular}{lrrrrrrrr}
\toprule
\textbf{Judge} & $n_{C1}$ & $n_{C2}$ & $n_{C3}$ & $n_{C4}$ & $n_{C5}$ & $n_{C6}$ & $n_{C7}$ & $n_{C8}$ \\
\midrule
GPT-4o            & 523 & 523 & 523 & 523 & 2092 & 2092 & 6276 & 100 \\
GPT-4o-mini       & 523 & 523 & 523 & 523 & 2092 & 2092 & 6276 & 100 \\
Claude Haiku 4.5  & 523 & 523 & 523 & 523 & 2091 & 2092 & 6274 & 100 \\
Claude Sonnet 4.5 & 523 & 523 & 523 & 523 & 2092 & 2092 & 6276 & 100 \\
Gemini 2.5 Flash  & 521 & 521 & 522 & 521 & 2053 & 2047 & 6139 & 100 \\
Gemini 2.5 Pro    & 522 & 523 & 521 & 523 & 2062 & 2057 & 6160 & 100 \\
Llama-3.3-70B     & 523 & 523 & 523 & 523 & 2092 & 2092 & 6276 & 100 \\
\bottomrule
\end{tabular}
}
\caption{Per-judge per-condition sample sizes on the \textsc{judgerEva-Standard} full set ($N_{\text{plain}}{=}2{,}092$; $C5{-}C7$ over all variants).}
\label{tab:sample_sizes_full}
\end{table}

Per-condition sample sizes on the \textsc{judgerEva-Standard} full set (Table~\ref{tab:sample_sizes_full}) reflect API and parse exclusions. Gemini 2.5 Flash and Gemini 2.5 Pro show $\sim2\%$ shortfall on $C5$--$C7$ from rate-limit retries that exhausted our retry budget. Llama-3.3-70B experienced a transient provider-side billing interruption (HTTP 402) on a contiguous block of items; those items were retried successfully in a follow-up pass and the final counts are at parity with the other judges. Per-item retry traces and parse-failure logs are released alongside the judge outputs.

\paragraph{Compute and cost.}
All judging, augmentation, and rationale evaluation reported in this paper run through commercial API endpoints; no local GPU training was performed and no model weights were updated. Across the seven judges, the experiment comprises ${\approx}87.6$k pairwise verdicts on \textsc{judgerEva-Standard} (plain, polished, and length-padded variants in both A/B and B/A orderings), ${\approx}1.9$k on \textsc{Hard-Constructed}, and ${\approx}39.8$k on the five external benchmarks (JudgeBench, JudgeBench-Pro, RewardBench, MT-Bench, LLMBar), for a total of approximately $129$k judge calls. The Claude Sonnet 4.6 rationale evaluator contributed an additional ${\approx}0.7$k calls on correct-verdict subsets. \textsc{judgerEva-Standard} construction (Appendix~\ref{app:judgereva}) and JudgeBench augmentation (Appendix~\ref{app:augmentation_prompts}) issued additional GPT-4o and Gemini 2.5 Pro calls for answer expansion, polishing, length padding, subtlety scoring, and Hard-Constructed hardening, with up to three regeneration retries gated by the subtlety filter. Experiments were orchestrated from a single laptop with no local accelerator; aggregate wall-clock time was dominated by provider rate limits rather than computation. Per-call records (request IDs, latencies, retry traces) are released alongside the judge outputs for exact replication of the compute budget.

\section{Estimators}
\label{app:estimators}

For completeness we restate the cross-benchmark and stress estimators referenced in \S\ref{sec:methodology}. Cross-benchmark consistency on a condition $c$ jointly measurable on benchmarks $\mathcal{D}$ and $\mathcal{D}'$ is
\begin{equation}
\rho_c(\mathcal{D},\mathcal{D}')=\mathrm{Spearman}\bigl(\mathbf{p}_c(\mathcal{D}),\mathbf{p}_c(\mathcal{D}')\bigr),
\end{equation}
where $\mathbf{p}_c(\mathcal{D})$ is the seven-judge vector of CAP values on $\mathcal{D}$. Adversarial stress degradation is
\begin{equation}
\label{eq:stress_degradation}
\delta^{\mathrm{str}}_c(j)=\mathrm{CAP}_c(j;\mathrm{JB})-\mathrm{CAP}_c(j;\mathrm{JB\mbox{-}Pro}),
\end{equation}
so positive values indicate worse performance under bias-injected stress. Within-family divergence is the $L_1$ distance between two judges' measurable CAP profiles on the same benchmark.

\section{Release Manifest and AI Use Statement}
\label{app:release}

The public release accompanying the camera-ready version includes: (i) the \textsc{judgerEva-Standard} benchmark (plain/polished/length-padded variants; Hard-Empirical and Hard-Constructed subsets); (ii) all generation, judging, and rationale-evaluation prompts; (iii) raw judge outputs for the seven judges on all six benchmarks, including swapped-order evaluations; (iv) JudgeBench augmentation outputs (polished and length-padded variants per judge); (v) the analysis pipeline producing the tables and figures in this paper, including the bootstrap-CI implementation; and (vi) a \texttt{compute\_cap.py} reference that produces a CAP profile on any pairwise judging dataset given per-item judge outputs and (optional) error-type annotations. The release is available at \url{https://github.com/wenqi-l/CAP-judgerEva}; code is MIT-licensed, and benchmark items are distributed for research use under the licenses of their original sources.

\paragraph{AI Use Statement}
AI systems were used in this work in three roles, each disclosed elsewhere in the paper. \emph{(i) As experimental subjects:} the seven judges and the rationale evaluator are the objects of study (\S\ref{sec:methodology}). \emph{(ii) Inside the data pipeline:} GPT-4o generated \textsc{judgerEva-Standard}'s flawed answers and length-padded variants, Gemini 2.5 Pro generated the Hard-Constructed hardening, and Claude Sonnet 4.6 served as the C8 rationale evaluator (Appendices~\ref{app:judgereva_response_generation},~\ref{app:judgereva_hard},~\ref{app:evaluator_prompt}); all prompts and per-item outputs are released. \emph{(iii) As authoring assistants:} AI coding assistants were used for implementation and debugging of the analysis pipeline, and large language models were used for sentence-level editing of authors' drafts (clarity, grammar, conciseness). All conceptual contributions, experimental design, claims, and conclusions are the authors'; the authors verified all AI-assisted code and reviewed all AI-edited prose before submission.

\end{document}